\documentclass[sigplan,,screen]{acmart}
\usepackage[utf8]{inputenc}

\usepackage[ruled,linesnumbered,vlined]{algorithm2e}
\usepackage[font=scriptsize, justification=justified,
   format=plain,skip=0pt,belowskip=0pt,aboveskip=4pt]{caption}
\usepackage{pifont}
\usepackage{color}
\usepackage{tabularray}
\usepackage{booktabs}
\usepackage{colortbl}
\usepackage{arydshln}
\usepackage{graphicx}
\graphicspath{{source/Figures/}}
\usepackage{multirow}
\usepackage{listings}

\usepackage[shortlabels,inline]{enumitem}
\usepackage{subcaption} 

\UseTblrLibrary{diagbox}
\usepackage{makecell}
\usepackage{hhline}

\SetAlgoSkip{}
\def\BibTeX{{\rm B\kern-.05em{\sc i\kern-.025em b}\kern-.08em
    T\kern-.1667em\lower.7ex\hbox{E}\kern-.125emX}}

\AtBeginDocument{%
  \providecommand\BibTeX{{%
    Bib\TeX}}}

\copyrightyear{2025}
\acmYear{2025}
\setcopyright{cc}
\setcctype{by}
\acmConference[UCC '25]{2025 IEEE/ACM 18th International Conference on Utility and Cloud Computing}{December 1--4, 2025}{France, France}
\acmBooktitle{2025 IEEE/ACM 18th International Conference on Utility and Cloud Computing (UCC '25), December 1--4, 2025, France, France}\acmDOI{10.1145/3773274.3774267}
\acmISBN{979-8-4007-2285-1/2025/12}

\begin{document}

\title{\texttt{Splitwise}: Collaborative Edge–Cloud Inference for LLMs via Lyapunov-Assisted DRL}

\author{Abolfazl Younesi}
\orcid{0009-0003-0052-6475}
\email{Abolfazl.Younesi@uibk.ac.at}
\affiliation{%
  \institution{University of Innsbruck}
  \city{Innsbruck}
  \country{Austria}
}

\author{Abbas Shabrang Maryan}
\orcid{0009-0003-0735-2455}
\email{abbas.shabrang81@sharif.edu}
\affiliation{%
  \institution{Sharif University of Technology}
  \city{Tehran}
  \country{Iran}
}

\author{Elyas Oustad}
\orcid{0009-0006-1456-356X}
\email{elyas.oustad23@sharif.edu}
\affiliation{%
  \institution{Sharif University of Technology}
  \city{Tehran}
  \country{Iran}
}

\author{Zahra Najafabadi Samani}
\orcid{0000-0001-5182-9087}
\email{Zahra.Najafabadi-Samani@uibk.ac.at}
\affiliation{%
  \institution{University of Innsbruck}
  \city{Innsbruck}
  \country{Austria}
}

\author{Mohsen Ansari}
\orcid{0000-0002-4670-8608}
\email{ansari@sharif.edu}
\affiliation{%
  \institution{Sharif University of Technology}
  \city{Tehran}
  \country{Iran}
}

\author{Thomas Fahringer}
\orcid{0000-0003-4293-1228}
\email{Thomas.Fahringer@uibk.ac.at}
\affiliation{%
  \institution{University of Innsbruck}
  \city{Innsbruck}
  \country{Austria}
}

\renewcommand{\shortauthors}{Younesi et al.}

\begin{abstract}
Deploying large language models (LLMs) on edge devices is challenging due to their limited memory and power resources. Cloud-only inference reduces device burden but introduces high latency and cost. Static edge–cloud partitions optimize a single metric and struggle when bandwidth fluctuates.
We propose \texttt{Splitwise}, a novel Lyapunov-assisted deep reinforcement learning (DRL) framework for fine-grained, adaptive partitioning of LLMs across edge and cloud environments. \texttt{Splitwise} decomposes transformer layers into attention heads and feed-forward sub-blocks, exposing exponentially more partition choices than layer-wise schemes. A hierarchical DRL policy, guided by Lyapunov optimization, jointly minimizes latency, energy consumption, and accuracy degradation while guaranteeing queue stability under stochastic workloads and variable network bandwidth. \texttt{Splitwise} also guarantees robustness via partition checkpoints with exponential backoff recovery in case of communication failures.
Experiments on Jetson Orin NX, Galaxy S23, and Raspberry Pi 5 with GPT‑2 (1.5B), LLaMA‑7B, and LLaMA‑13B show that \texttt{Splitwise} reduces end‑to‑end latency by 1.4$\times$–2.8$\times$ and cuts energy consumption by up to 41\% compared with existing partitioners. It lowers the 95th-percentile latency by 53–61\% relative to cloud-only execution, while maintaining accuracy and modest memory requirements.
\end{abstract}



\maketitle

\section{Introduction} \label{sec:introduction}
The deployment of Large Language Models (LLMs) has revolutionized numerous applications, from intelligent assistants to code generation \cite{zhang2024edgeshard,jin2024collm,boateng2025survey}. However, their computational demands often exceed billions of parameters, posing significant challenges for resource-constrained edge devices. While cloud-based inference offers abundant computational resources, it introduces substantial latency (50-200ms), cost, and raises privacy concerns for sensitive applications \cite{he2024large,li2025collaborative,ye2025jupiter}.
This latency is prohibitive for the next generation of interactive edge applications. For example, real-time conversational agents and augmented reality (AR) assistants all require a "perceived-as-instant" response (ideally sub-100ms) to maintain a fluid user experience. A 50-200ms network round-trip makes this target impossible before inference computation even begins. Furthermore, many use cases involve processing sensitive data, such as private messages and medical transcriptions, where offloading to the cloud is non-viable due to privacy constraints. These applications must run at least partially on-device, creating the exact resource bottleneck that collaborative inference aims to solve.
Edge-only deployment, conversely, suffers from limited memory (4-8GB) and computational capacity, resulting in either model quality degradation through aggressive compression or prohibitive energy consumption \cite{pan2025instattention,noh2025adaptive}.

This gap highlights a collaborative edge-cloud inference paradigm that strategically utilizes both edge proximity and cloud capacity \cite{zhu2025lslm,li2025collaborative,noh2025adaptive,8457785}.
These scenarios frequently exhibit rapid variations in both wireless bandwidth (10–100 Mbps) and bursty request arrivals caused by interactive user behavior. Under such dynamics, a static partition quickly becomes suboptimal or even unstable, leading to queue buildup and tail latency violations. Thus, dynamic partitioning is essential for maintaining responsiveness and resource efficiency in practical deployments of LLM-based services.

\textbf{Limitations of existing works.} Current approaches to edge-cloud collaborative inference exhibit three fundamental limitations. First, static partitioning strategies~\cite{10.1145/3037697.3037698,Laskaridis_2020,10.1007/3-540-57659-2_11} pre-determine the split point between edge and cloud, failing to adapt to dynamic network conditions and varying workload characteristics. For instance, when network bandwidth fluctuates from 10 to 100 Mbps, which is common in mobile scenarios, static approaches either underutilize available bandwidth or suffer from severe bottlenecks~\cite{zhang2024edgeshard,li2019edgeaiondemandaccelerating}. Second, existing methods typically partition models at coarse granularities (e.g., entire transformer layers), missing opportunities for fine-grained optimization~\cite{8737614,li2025collaborative}. This coarse partitioning limits the ability to optimize resource allocation at a sub-layer level, reducing efficiency in heterogeneous edge environments~\cite{zhang2025tensallo}. Third, prior work often optimizes for single objectives, such as latency or energy, overlooking the complex interplay between latency, energy consumption, and model accuracy that characterizes real-world deployments~\cite{he2024large,chen2025adaptive,tian2025clone}. This single-objective focus fails to address the multi-dimensional trade-offs required for practical edge-cloud systems~\cite{narayan2025minions}.

\textbf{Proposed approach.} We present \texttt{Splitwise}, a novel framework that dynamically partitions LLMs between edge devices and cloud servers using Lyapunov-assisted Deep Reinforcement Learning (DRL). Unlike existing approaches, \texttt{Splitwise} formulates the partitioning problem as a constrained Markov Decision Process (MDP) where actions determine fine-grained partition points at the attention head and FFN sub-layer level. The Lyapunov optimization framework provides theoretical guarantees on queue stability, ensuring bounded latency even under stochastic arrivals and time-varying network conditions. Our DRL agent learns to balance immediate performance metrics (latency, energy) with long-term system stability by incorporating Lyapunov drift into the reward function. 

\begin{figure}[!t]
    \centering
    \begin{subfigure}[t]{0.49\columnwidth}
        \centering
            \includegraphics[width=1\linewidth]{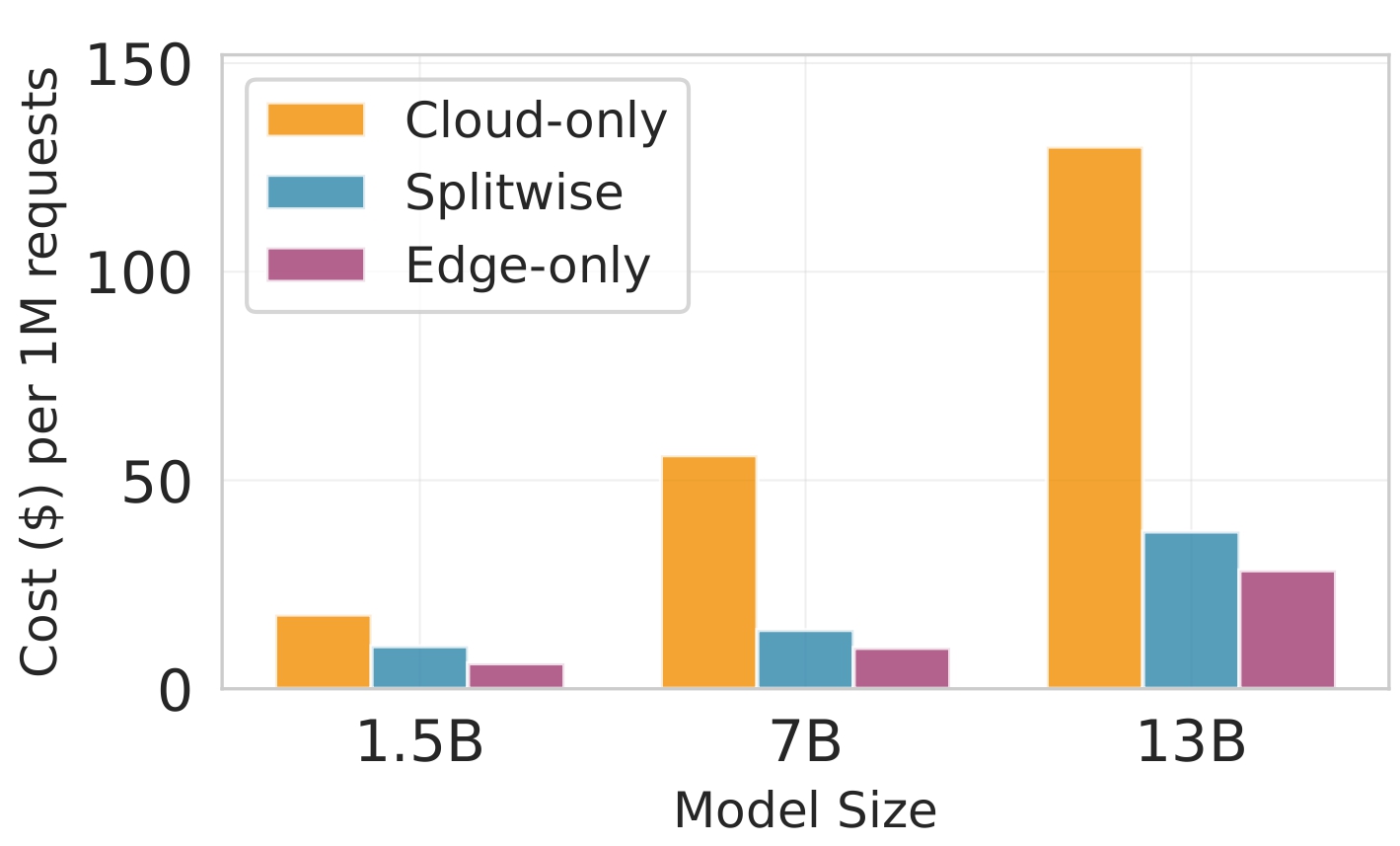}
        \caption{Operational Cost Analysis}
        \label{fig:Operational Cost}
    \end{subfigure}
    \begin{subfigure}[t]{0.49\columnwidth}
        \centering
            \includegraphics[width=1\linewidth]{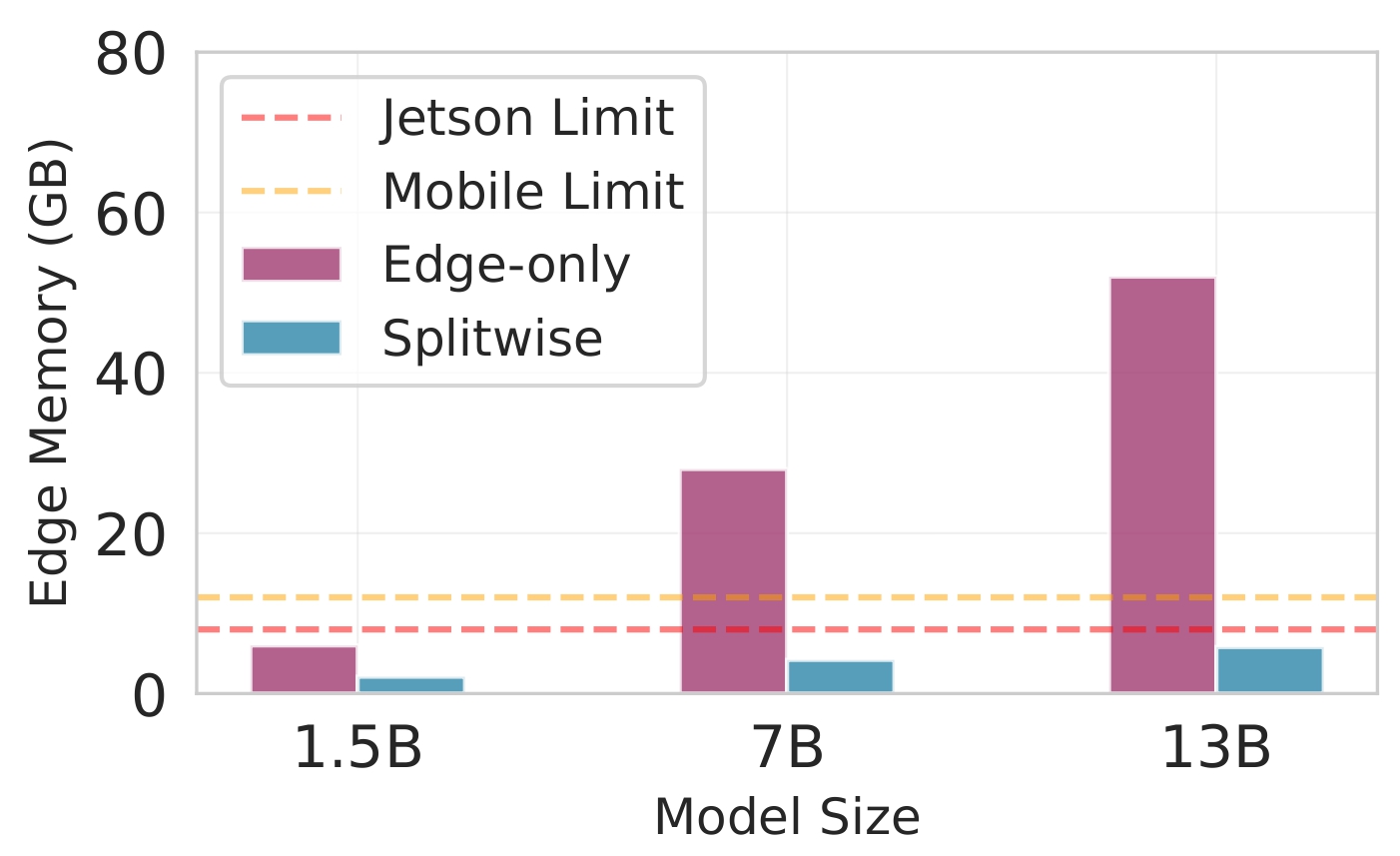}
        \caption{Memory Requirements}
        \label{fig:Memory}
    \end{subfigure}

    \begin{subfigure}[t]{0.8\columnwidth}
        \centering
            \includegraphics[width=1\linewidth]{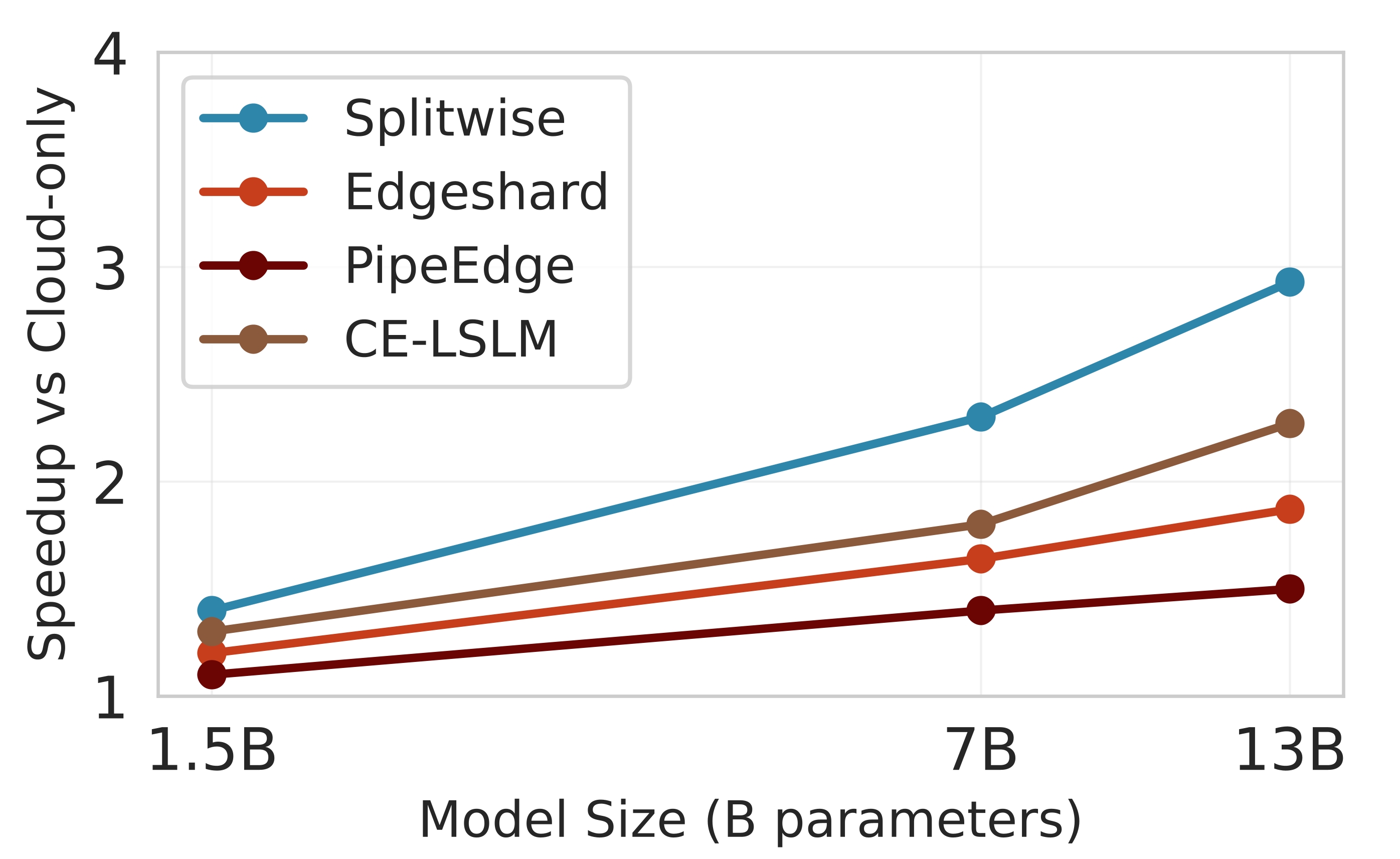}
        \caption{performance scaling compared to cloud-only execution}
        \label{fig:Performance Scaling}
    \end{subfigure}

    \caption{Illustration of key experimental metrics, highlighting variations in (a) operational cost, (b) memory requirements, and (c) performance scaling compared to cloud-only execution. (Details of setup are in Section \ref{sec:performance})}
    \label{fig:motive}
\end{figure}

We demonstrate some critical tradeoffs that define the edge-cloud inference landscape for our \texttt{\texttt{Splitwise}} based on our observations in Figure~\ref{fig:motive}.
Figure~\ref{fig:Operational Cost} highlights the operational costs of various inference strategies. As model size grows from 1.5 billion (B) to 13B parameters, the cost per million requests for cloud-only execution increases dramatically, surpassing \$150 for the largest model. This is due to the need for robust cloud infrastructure. In contrast, edge-only execution remains cost-effective but is limited by memory constraints for larger models. \texttt{\texttt{Splitwise}} significantly reduces costs, keeping them below \$25 (6$\times$ lower than cloud only) even for the 13B parameter model by utilizing edge and cloud resources.
Figure~\ref{fig:Memory} shows the memory footprint on edge devices. The 13B parameter model requires over 50 GB, exceeding the 8 GB limit of typical mobile devices and the 16 GB limit of most edge gateways, making large-scale inference difficult. \texttt{\texttt{Splitwise}} addresses this by reducing the required edge memory to 10 GB through effective computation offloading.
Finally, Figure~\ref{fig:Performance Scaling} illustrates the performance gains from collaborative inference, achieving up to a 3$\times$ speedup for the 13B parameter model. 
Overall, these results underscore the advantages of \texttt{\texttt{Splitwise}}. They show that traditional approaches are often too costly and inefficient.

\textbf{Insights.} We have two key insights based on our observations. First, the computational cost and data transfer requirements vary significantly across different components of LLMs. Attention mechanisms in LLMs require quadratic computation but produce compact representations, while feed-forward networks (FFNs) exhibit opposite characteristics \cite{li2025collaborative,pan2025instattention}. This heterogeneity suggests that optimal partition points differ based on runtime conditions. Second, edge-cloud systems inherently exhibit queue dynamics, where request arrivals and processing rates create complex stability challenges that pure optimization approaches fail to address.

\textbf{Contributions.} This paper makes the following contributions:
\begin{itemize}[wide]
\item[\textbf{C1)}] To the best of our knowledge, we propose the first Lyapunov-assisted DRL framework called \texttt{Splitwise} for dynamic LLM partitioning that guarantees system stability while jointly optimizing latency and energy consumption in the edge-cloud collaborative inference (see Section \ref{sec:architecture}).
\item[\textbf{C2)}] We provide a fault-tolerance method that uses partition-boundary checkpoints to recover from communication failures, resuming execution with exponential backoff.
\item[\textbf{C3)}] We introduce a hierarchical partitioning scheme that operates at the level of attention heads and feed-forward sub-blocks, exposing hundreds of times more partition points than traditional layer-wise approaches (see Section \ref{subsec:edge-cloud}). This fine granularity enables \texttt{Splitwise} to adapt effectively under fluctuating bandwidth and workload conditions.
\item[\textbf{C4)}] We demonstrate through extensive evaluation on real edge devices (NVIDIA Jetson, mobile phones) and various LLMs (1B-13B parameters) that \texttt{Splitwise} reduces latency by 1.4$\times$–2.8$\times$  while saving energy by up to 41\% compared to state-of-the-art baselines. Moreover, \texttt{Splitwise} reduces operational inference costs by up to 6$\times$ compared to a cloud-only deployment (see Section \ref{sec:performance}).

\end{itemize}

\textbf{Paper structure.} The remainder of this paper is organized as follows.  Section II presents our system model and problem formulation. Section III details the Lyapunov-assisted DRL framework. Section IV describes implementation optimizations. Section V evaluates \texttt{Splitwise}. Section VI reviews related work. Section VII concludes the paper.

\section{System Model and Problem Formulation} \label{sec:background}
We first present our system model for edge-cloud collaborative LLM inference, followed by the problem formulation. Table~\ref{tab:notations} summarizes the key notations used in this paper.

\subsection{Edge-Cloud Architecture}
\label{subsec:edge-cloud}
\textbf{System components.} We consider an edge–cloud collaborative system consisting of a set of edge devices $\mathcal{E} = \{E_1, E_2, \dots, E_{N_e}\}$, each with limited computational resources, and a set of cloud servers $\mathcal{C} = \{C_1, C_2, \dots, C_{N_c}\}$ with abundant capacity. Each edge device $E_i \in \mathcal{E}$ is characterized by its compute capability $CC_e^i$ (FLOPs), memory capacity $M_e^i$ (GB), and power budget $P_e^i$ (W). The cloud servers provide substantially higher aggregate resources, i.e., $CC_c^j \gg CC_e^i$ for most $j,i$, but are accessible only through network links with time-varying bandwidth $B_i(t)$ and latency $l_{n,i}(t)$ associated with each edge–cloud connection \cite{zhang2024edgeshard}.

\begin{table}
\centering
\setlength{\extrarowheight}{0pt}
\setlength{\aboverulesep}{0pt}
\setlength{\belowrulesep}{0pt}
\caption{Summary of key notations used throughout the paper}
\label{tab:notations}
\resizebox{\linewidth}{!}{%
\begin{tabular}{ll} 
\toprule
\rowcolor[rgb]{0.816,0.816,0.816} \textbf{Notation} & \textbf{Description} \\ 
\hline
\multicolumn{2}{l}{\textit{System Architecture}} \\
$\mathcal{E}, \mathcal{C}$ & Edge device and cloud server \\
$CC_e, CC_c$, $M_e$, $P_e$ & Compute (FLOPS), memory (GB), and power (W) of edge/cloud \\
$B(t)$, $l_n(t)$ & Network bandwidth and latency at time $t$ \\
\hline
\multicolumn{2}{l}{\textit{Model Architecture}} \\
$L, H$ & Number of transformer layers and~ attention heads per layer \\
$d_{model}$ & Model hidden dimension \\
$d_h$ & Attention head dimension ($d_{model}/H$) \\
$d_{ff}$ & Feed-forward network dimension \\
$\theta_i$ & Parameters of layer $i$ \\
$X^{(\ell)}$ & Input tensor to layer $\ell$ \\ 
\hline
\multicolumn{2}{l}{\textit{Partitioning}} \\
$\pi$ & Complete partitioning strategy across all layers \\
$\pi^{(\ell)}$ & Partition decision for layer $\ell$ \\
$\pi_{h}^{(\ell)}$ & Placement of head $h$ in layer $\ell$ (0: edge, 1: cloud) \\
$\pi_{FFN}^{(\ell)}$ & FFN partition mode (0: edge, 1: cloud, 2: split) \\
$\mathcal{E}_{h}^{(\ell)}$, $\mathcal{C}_{h}^{(\ell)}$ & Set of heads assigned to edge and cloud in layer $\ell$ \\
$\mathcal{B}(\pi)$ & Set of partition boundaries \\ 
\hline
\multicolumn{2}{l}{\textit{Performance Metrics}} \\
$T(\pi, t)$ & End-to-end inference latency under partition $\pi$ \\
$T_{comp}^e$, $T_{comp}^c$ & Computation time on edge and cloud \\
$T_{comm(\pi, t)}$, $E(\pi)$ & Communication time and Energy consumption for partition $\pi$ \\
$\Delta_{acc(\pi)}$ & Accuracy degradation from partitioning \\
$Q_{b(\cdot)}$ & Quantization function at boundary $b$ \\ 
\hline
\multicolumn{2}{l}{\textit{Queue Dynamics}} \\
$Q(t)$ & Queue backlog at time $t$ \\
$\lambda(t)$,$\mu(\pi, t)$ & Request arrival rate,~Service rate under partition~$\pi$ \\
$A(t)$ & New arrivals in time slot $t$ \\
$L(Q(t))$, $\Delta(Q(t))$ & Lyapunov function,~Conditional Lyapunov drift \\ 
\hline
\multicolumn{2}{l}{\textit{Reinforcement Learning}} \\
$s_t$, $a_t$ & System state and action (partition decision) at time $t$ \\
$r(s_t, a_t)$ & Reward function \\
$\pi_\theta$ & Policy network with parameters $\theta$ \\
$V_\phi^{perf}$, $V_\psi^{stab}$ & Performance and Stability critic network \\
$V$ & Lyapunov control parameter \\
$g(s_t, a_t)$ & Immediate cost function \\
$\gamma$ & Discount factor \\
$\tau_{temp}$ & Temperature for Gumbel-softmax \\
\bottomrule
\end{tabular}
}
\end{table}
\textbf{LLM architecture.} We consider transformer-based LLMs consisting of $L$ sequential layers. Each layer $\ell \in \{1,...,L\}$ processes input tensor $X^{(\ell)} \in \mathbb{R}^{n \times d_{model}}$ where $n$ denotes sequence length and $d_{model}$ is the model dimension. The layer computation follows:
\begin{equation}
X^{(\ell+1)} = \text{FFN}^{(\ell)}(\text{MHA}^{(\ell)}(X^{(\ell)}) + X^{(\ell)}) + \text{MHA}^{(\ell)}(X^{(\ell)}) + X^{(\ell)}
\end{equation}
The Multi-Head Attention (MHA) module decomposes into $H$ parallel attention heads:
\begin{equation}
\text{MHA}^{(\ell)}(X) = \text{Concat}(\text{head}_1^{(\ell)}, ..., \text{head}_H^{(\ell)})W_O^{(\ell)}
\end{equation}
where each head $h \in \{1,...,H\}$ independently computes:
\begin{equation}
\text{head}_h^{(\ell)} = \text{Attention}(XW_{Q,h}^{(\ell)}, XW_{K,h}^{(\ell)}, XW_{V,h}^{(\ell)})
\end{equation}

with projection matrices $W_{Q,h}^{(\ell)}, W_{K,h}^{(\ell)}, W_{V,h}^{(\ell)} \in \mathbb{R}^{d_{model} \times d_h}$ where $d_h = d_{model}/H$. The Feed-Forward Network (FFN) consists of two linear transformations with activation:
\begin{equation}
\text{FFN}^{(\ell)}(X) = \max(0, XW_1^{(\ell)} + b_1^{(\ell)})W_2^{(\ell)} + b_2^{(\ell)}
\end{equation}

where $W_1^{(\ell)} \in \mathbb{R}^{d_{model} \times d_{ff}}$ and $W_2^{(\ell)} \in \mathbb{R}^{d_{ff} \times d_{model}}$ with intermediate dimension $d_{ff} = 4d_{model}$ typically.

\textbf{Partitioning granularity.} We introduce a hierarchical partitioning scheme that enables flexible distribution of computation. Let $\pi = \{\pi^{(1)}, ..., \pi^{(L)}\}$ denote the complete partitioning strategy, where each layer's partition $\pi^{(\ell)}$ is defined as:
\begin{equation}
\pi^{(\ell)} = (\pi_{MHA}^{(\ell)}, \pi_{FFN}^{(\ell)})
\end{equation}
The MHA partition $\pi_{MHA}^{(\ell)} = [\pi_{h_1}^{(\ell)}, ..., \pi_{h_H}^{(\ell)}] \in \{0,1\}^H$ specifies the placement of each attention head, where $\pi_{h_i}^{(\ell)} = 0$ indicates edge execution and $\pi_{h_i}^{(\ell)} = 1$ indicates cloud execution. The FFN partition $\pi_{FFN}^{(\ell)} \in \{0, 1, 2\}$ supports three modes:
\begin{itemize}[leftmargin=*]
\item $\pi_{FFN}^{(\ell)} = 0$: Entire FFN executes on edge
\item $\pi_{FFN}^{(\ell)} = 1$: Entire FFN executes on cloud  
\item $\pi_{FFN}^{(\ell)} = 2$: Split execution with $W_1^{(\ell)}$ on edge and $W_2^{(\ell)}$ on cloud
\end{itemize}

This formulation enables $2^H \times 3$ possible configurations per layer, yielding a total action space of $(2^H \times 3)^L$ partitions. For a 24-layer model with 16 heads, this creates approximately $10^{31}$ possible configurations, necessitating intelligent exploration strategies \cite{kafetzis2025large}.

\textbf{Data flow formalization.} Given partition $\pi^{(\ell)}$, the data flow through layer $\ell$ involves potential edge-cloud transitions. Let $\mathcal{E}_h^{(\ell)} = \{h : \pi_{h}^{(\ell)} = 0\}$ and $\mathcal{C}_h^{(\ell)} = \{h : \pi_{h}^{(\ell)} = 1\}$ denote edge and cloud head assignments \cite{zhang2024edgeshard}. The computation proceeds as:
\begin{align}
Y_{edge}^{(\ell)} &= \sum_{h \in \mathcal{E}_h^{(\ell)}} \text{head}_h^{(\ell)}(X^{(\ell)})W_{O,h}^{(\ell)} \\
Y_{cloud}^{(\ell)} &= \sum_{h \in \mathcal{C}_h^{(\ell)}} \text{head}_h^{(\ell)}(X^{(\ell)})W_{O,h}^{(\ell)}
\end{align}
The aggregation requires communication if $|\mathcal{E}_h^{(\ell)}| > 0$ and $|\mathcal{C}_h^{(\ell)}| > 0$, transferring partial results of size $\mathcal{O}(n \times d_{model})$ across the network \cite{zhang2024edgeshard}.

\subsection{Performance Metrics}
\textbf{Latency model.} The end-to-end inference latency $T(\pi, t)$ for partition $\pi$ at time $t$ comprises three components:
\begin{equation}
T(\pi, t) = T_{comp}^e(\pi) + T_{comm}(\pi, t) + T_{comp}^c(\pi)
\end{equation}
where $T_{comp}^e(\pi) = \sum_{i=1}^{L} \sum_{j \in \mathcal{H}_i^e} \frac{F_{ij}}{C_e}$ represents edge computation time for heads $\mathcal{H}_i^e$ assigned to edge with $F_{ij}$ FLOPs, $T_{comm}(\pi, t) = \sum_{k=1}^{K(\pi)} \frac{D_k(\pi)}{B(t)} + l_n(t)$ captures communication overhead for $K(\pi)$ edge-cloud transitions with data volume $D_k(\pi)$, and $T_{comp}^c(\pi)$ denotes cloud computation time.

\textbf{Energy consumption.} The edge device energy consumption combines computation and communication costs \cite{yuan2024generative}:
\begin{equation}
E(\pi) = \sum_{i=1}^{L} \sum_{j \in \mathcal{H}_i^e} P_e^{comp} \cdot t_{ij} + \sum_{k=1}^{K(\pi)} P_e^{comm} \cdot \frac{D_k(\pi)}{B(t)}
\end{equation}
where $P_e^{comp}$ and $P_e^{comm}$ denote power consumption for computation and communication respectively, and $t_{ij}$ is the execution time for component $j$ in layer $i$ \cite{yuan2024generative}.

\textbf{Accuracy preservation.} Partitioning introduces quantization at boundaries to reduce communication. Let $Q_b$ denote the quantization function at boundary $b$. The accuracy degradation is modeled as:
\begin{equation}
\Delta_{acc}(\pi) = \sum_{b \in \mathcal{B}(\pi)} \alpha_b \cdot \|x_b - Q_b(x_b)\|_2^2
\end{equation}
where $\mathcal{B}(\pi)$ represents partition boundaries, $x_b$ is the activation tensor at boundary $b$, and $\alpha_b$ weights the importance of each boundary based on gradient flow analysis.

\subsection{Queue Dynamics and Stability}
\textbf{Request queue model.} Inference requests arrive according to a stochastic process with rate $\lambda(t)$. We model the queue backlog $Q(t)$, representing the number of unprocessed requests at time $t$. The queue evolution follows:
\begin{equation}
Q(t+1) = \max[Q(t) - \mu(\pi, t), 0] + A(t)
\end{equation}
where $\mu(\pi, t) = 1/T(\pi, t)$ is the service rate under partition $\pi$, and $A(t)$ represents new arrivals in slot $t$ \cite{bae2020reinforcement}.

\textbf{Lyapunov function.} To ensure queue stability, we define the Lyapunov function:
\begin{equation}
L(Q(t)) = \frac{1}{2}Q(t)^2
\end{equation}
The conditional Lyapunov drift $\Delta(Q(t))$ measures expected change in queue backlog:
\begin{equation}
\Delta(Q(t)) = \mathbb{E}[L(Q(t+1)) - L(Q(t)) | Q(t)]
\end{equation}

\subsection{Problem Formulation}
\textbf{Objective function.} We formulate the dynamic partitioning problem as minimizing a weighted combination of latency, energy, and accuracy loss while maintaining queue stability:
\begin{equation}\footnotesize
\min_{\pi(t)} \mathbb{E}\left[\sum_{t=0}^{\infty} \gamma^t \left(w_T T(\pi(t), t) + w_E E(\pi(t)) + w_A \Delta_{acc}(\pi(t))\right)\right]
\end{equation}
subject to:
\begin{align}
& \limsup_{t \to \infty} \frac{1}{t} \sum_{\tau=0}^{t-1} \mathbb{E}[Q(\tau)] < \infty \quad \text{(queue stability)} \\
& \sum_{j \in \mathcal{H}_i^e} m_{ij} \leq M_e, \quad \forall i \quad \text{(memory constraint)} \\
& E(\pi(t)) \leq P_e \cdot T(\pi(t), t) \quad \text{(power constraint)}
\end{align}

where $\gamma \in (0,1)$ is the discount factor, $w_T$, $w_E$, $w_A$ are importance weights, and $m_{ij}$ denotes memory requirement for component $j$ in layer $i$.

\textbf{Constrained MDP formulation.} We cast this optimization as a constrained Markov Decision Process (MDP) with:
\begin{itemize}[leftmargin=*]
\item \textit{State space} $\mathcal{S}$: $s_t = (Q(t), B(t), \lambda(t), \mathcal{R}(t))$ capturing queue backlog, network bandwidth, arrival rate, and resource availability (see details in Section \ref{sec:state-action})
\item \textit{Action space} $\mathcal{A}$: $a_t = \pi(t)$ representing partitioning decisions (see details in Section \ref{sec:state-action})
\item \textit{Transition dynamics}: $P(s_{t+1}|s_t, a_t)$ governed by queue evolution and stochastic network/workload changes 
\item \textit{Reward function}: Combining immediate cost and Lyapunov drift as detailed in Section \ref{sec:reward}
\end{itemize}

The challenge lies in solving this constrained MDP with continuous state space, exponentially large action space, and stability requirements, which we address through our Lyapunov-assisted RL framework.

\section{\texttt{Splitwise} Design Overview} \label{sec:architecture}
\begin{figure}[!]
    \centering
    \begin{subfigure}[t]{0.49\columnwidth}
        \centering
            \includegraphics[width=1\linewidth]{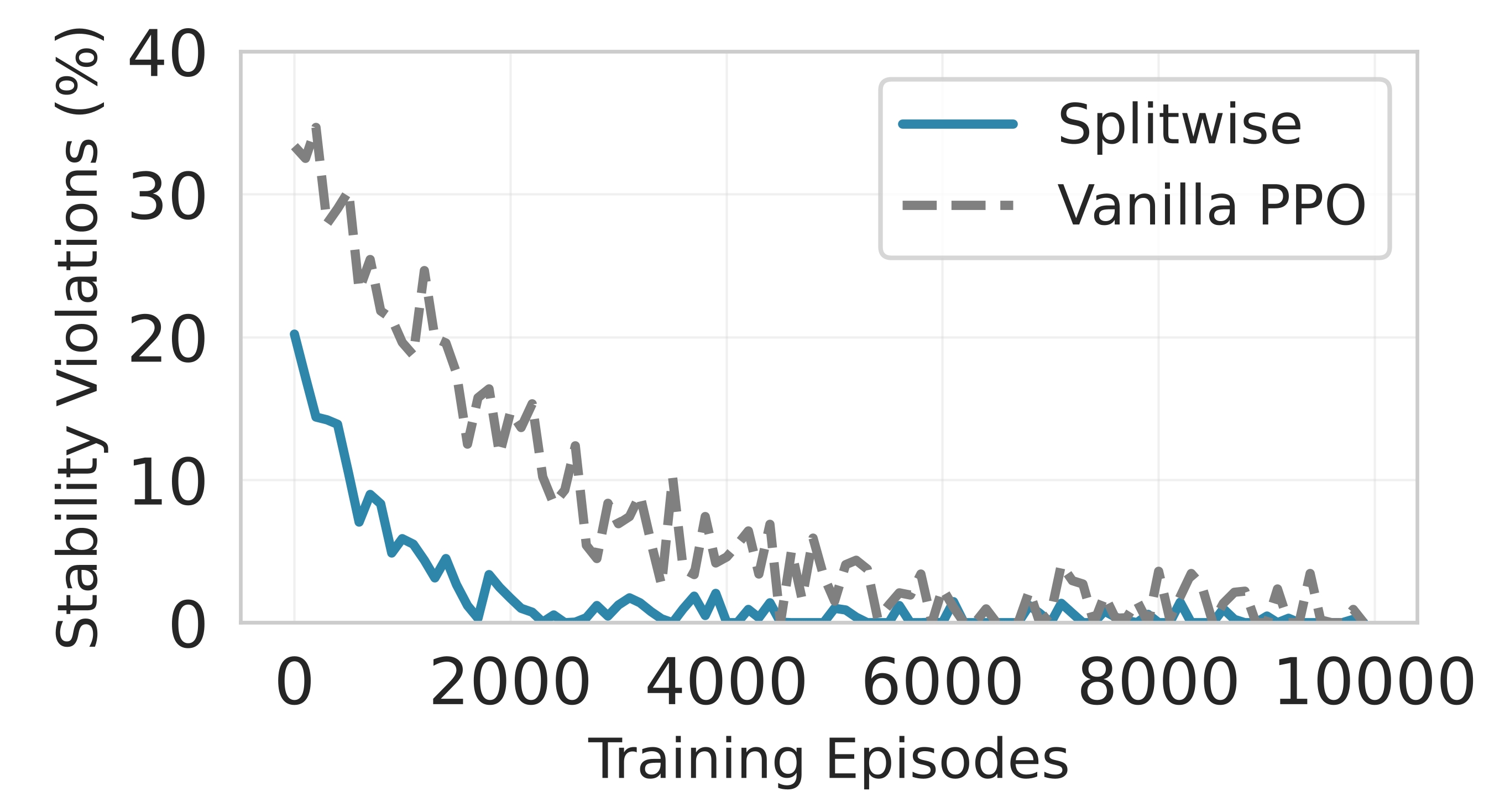}
        \vspace{-0.6cm}
        \caption{Training Stability}
        \label{fig:Training Stability}
    \end{subfigure}
    \begin{subfigure}[t]{0.49\columnwidth}
        \centering
            \includegraphics[width=1\linewidth]{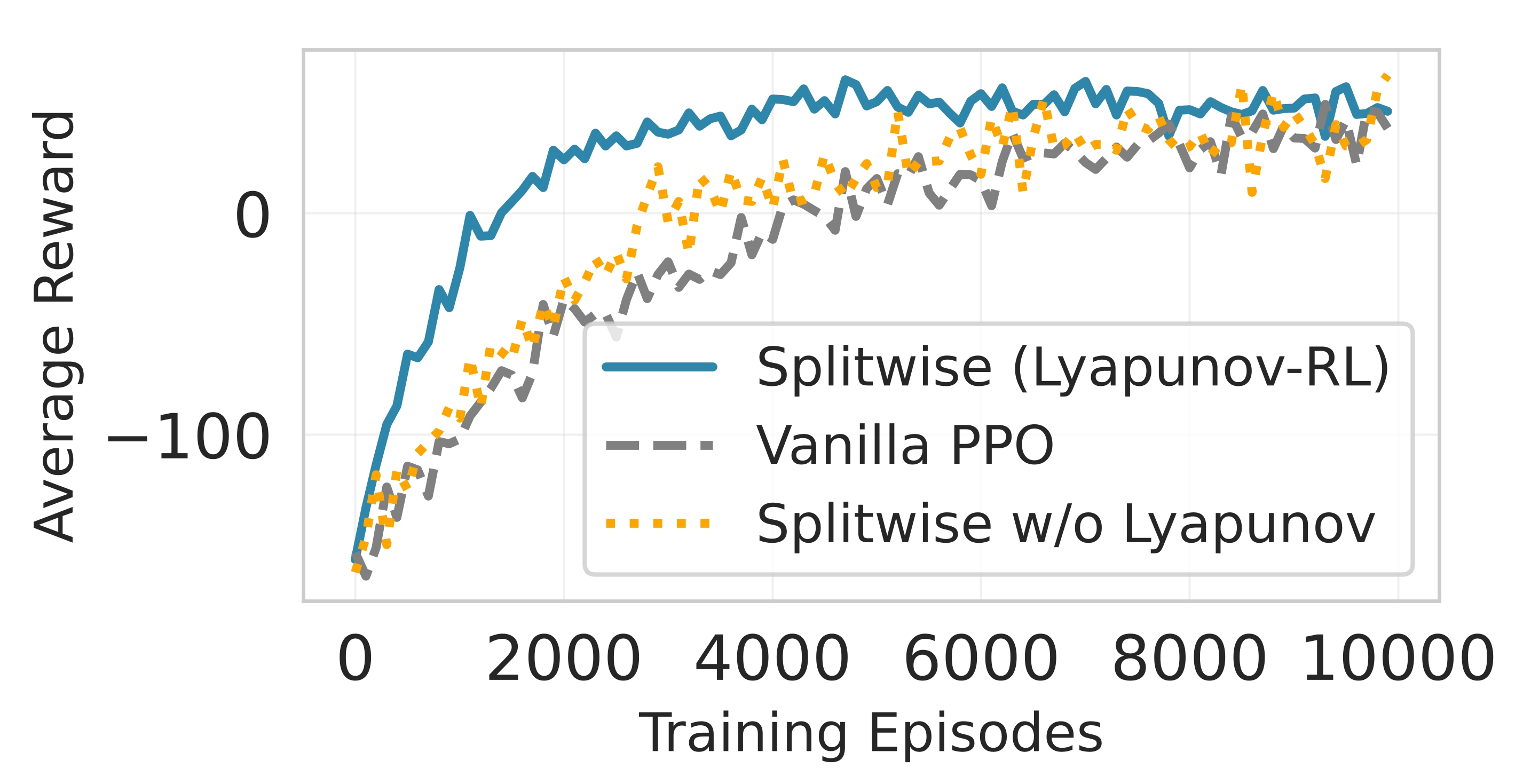}
        \vspace{-0.6cm}
        \caption{Learning Efficiency}
        \label{fig:Learning Efficiency}
    \end{subfigure}

    \caption{Comparison of model training dynamics, showing (a) training stability and (b) learning efficiency across different settings.}
    \label{fig:RL}
\end{figure}

\texttt{Splitwise} framework combines deep reinforcement learning with Lyapunov optimization theory to achieve both optimal performance and guaranteed stability. The framework consists of three key components: (i) a policy network $\pi_\theta$ that learns partitioning decisions, (ii) a Lyapunov critic that evaluates long-term stability, and (iii) a drift-plus-penalty reward function that balances immediate performance with queue stability. Unlike standard RL methods that may converge to unstable policies, \texttt{Splitwise} explicitly incorporates stability constraints into the learning process (cf. Fig. \ref{fig:RL}).

\subsection{State and Action Representation} \label{sec:state-action}

\textbf{State encoding.} We encode the system state $s_t$ as a comprehensive feature vector capturing both instantaneous conditions and temporal dynamics:
\begin{equation}\small
s_t = [\underbrace{Q(t), \bar{Q}_\tau}_{\text{queue}}, \underbrace{B(t), \bar{B}_\tau, \sigma_B}_{\text{network}}, \underbrace{\lambda(t), \bar{\lambda}_\tau}_{\text{workload}}, \underbrace{C_e^{avail}, M_e^{avail}}_{\text{resources}}, \underbrace{h_t}_{\text{history}}]
\end{equation}
where $\bar{Q}_\tau = \frac{1}{\tau}\sum_{i=t-\tau}^{t} Q(i)$ represents the moving average queue length over window $\tau$, $\sigma_B$ captures network bandwidth variance, and $h_t \in \mathbb{R}^{d_h}$ is a learned history embedding from an LSTM that encodes past partitioning decisions and their outcomes \cite{ha2018world}.

\textbf{Hierarchical action decomposition.} To handle the exponentially large action space $(2^H \times 3)^L$, we decompose actions hierarchically. Instead of selecting from all possible partitions, we structure the action as:
\begin{equation}
a_t = \pi_\theta(s_t) = [\alpha^{(1)}, ..., \alpha^{(L)}]
\end{equation}
where each layer action $\alpha^{(\ell)}$ is generated by:
\begin{equation}
\alpha^{(\ell)} = \text{softmax}(f_\theta^{(\ell)}(s_t, e^{(\ell-1)}))
\end{equation}
Here, $f_\theta^{(\ell)}$ is a layer-specific sub-network and $e^{(\ell-1)}$ encodes decisions from previous layers to capture inter-layer dependencies. Each $\alpha^{(\ell)} \in [0,1]^{H+1}$ represents continuous probabilities for placing each attention head and FFN on the cloud, which are discretized during execution using Gumbel-softmax:
\begin{equation}
\pi_{h}^{(\ell)} = \text{Gumbel-Softmax}(\alpha_h^{(\ell)}, \tau_{temp})
\end{equation}

where $\tau_{temp}$ is a temperature parameter annealed during training to transition from exploration to exploitation.

\subsection{Lyapunov-Guided Reward Design} \label{sec:reward}

\textbf{Drift-plus-penalty formulation.} Traditional RL optimizes expected cumulative reward without stability guarantees. We incorporate Lyapunov drift to ensure queue stability while optimizing performance. The reward function combines immediate cost with drift penalty:
\begin{equation}
r(s_t, a_t) = -[V \cdot \Delta(Q(t)) + g(s_t, a_t)]
\end{equation}

where $V > 0$ is a control parameter and $g(s_t, a_t)$ represents the immediate cost:
\begin{equation}
g(s_t, a_t) = w_T T(\pi_t, t) + w_E E(\pi_t) + w_A \Delta_{acc}(\pi_t)
\end{equation}

\textbf{Lyapunov drift computation.} The one-step Lyapunov drift under action $a_t$ is:
\begin{equation}
\begin{aligned}
\Delta(Q(t)) &= \mathbb{E}[L(Q(t+1)) - L(Q(t)) \mid Q(t), a_t] \\
&\leq B + Q(t)\,\mathbb{E}[A(t) - \mu(\pi_t, t)]
\end{aligned}
\end{equation}
where $B$ is a finite constant bounding the second moment of arrivals and service. This upper bound provides a tractable optimization target while maintaining guarantees.

\textbf{Adaptive weight adjustment.} The control parameter $V$ balances performance optimization against stability. We adaptively adjust $V$ based on the queue backlog:
\begin{equation}
V(t) = V_{min} + (V_{max} - V_{min}) \cdot \exp\left(-\frac{Q(t)}{Q_{ref}}\right)
\end{equation}

where $Q_{ref}$ is a reference queue length. This ensures aggressive performance optimization when queues are stable, while prioritizing stability when the backlog increases.

\subsection{Policy Learning Algorithm} \label{sec:policy}

\textbf{Actor-critic architecture.} We employ a Proximal Policy Optimization (PPO) algorithm \cite{schulman2017proximal} with dual critics to separately evaluate performance and stability:

\begin{itemize}[leftmargin=*,wide]
\item \textit{Performance critic} $V_\phi^{perf}(s)$: Estimates expected cumulative performance cost
\item \textit{Stability critic} $V_\psi^{stab}(s)$: Estimates expected queue backlog evolution
\end{itemize}

The combined value function is:
\begin{equation}
V(s) = V_\phi^{perf}(s) + V \cdot V_\psi^{stab}(s)
\end{equation}

\textbf{Policy gradient with stability constraints.} The policy gradient incorporates both performance and stability objectives:
\begin{equation}
\nabla_\theta J = \mathbb{E}_{\tau \sim \pi_\theta}\left[\sum_t \nabla_\theta \log \pi_\theta(a_t|s_t) A_t^{total}\right]
\end{equation}

where the total advantage function combines both critics:
\begin{equation}
A_t^{total} = A_t^{perf} + V \cdot A_t^{stab}
\end{equation}

with advantages computed using Generalized Advantage Estimation (GAE) for variance reduction.

\textbf{Algorithm overview.} Our training procedure combines Lyapunov optimization with deep reinforcement learning to jointly optimize performance and guarantee stability. Algorithm 1 presents the core training loop with detailed explanations of each component.
\begin{algorithm}[h]
\caption{Lyapunov-Assisted DRL Training}
\SetAlgoLined
\SetAlgoNoEnd
\scriptsize
\KwIn{Policy network $\pi_\theta$, critics $V_\phi^{perf}$, $V_\psi^{stab}$}
\KwOut{Optimized policy $\pi_\theta^*$}
\textbf{Initialize:} $\pi_\theta$, critics $V_\phi^{perf}$, $V_\psi^{stab}$, $\mathcal{D} \leftarrow \emptyset$,  $Q(0) = 0$, $\tau_{temp} = 1.0$, $V = V_{init}$\;
\For{episode $k = 1, 2, ..., K$}{ \label{line:episode_start}
   $s_0, \tau_k \leftarrow \textsc{GetInitialState}(), \emptyset$\;
   \For{step $t = 0, 1, ..., T_{max}$}{ \label{line:step_start}
       $a_t \sim \pi_\theta(a|s_t)$ \tcp*{Sample action from policy}
       $s_{t+1}, c_t \leftarrow$ \textsc{ExecutePartition}$(a_t)$ \tcp*{Execute partition} \label{line:execute}
       \tcp{Compute Lyapunov drift}
       $\mu_t \leftarrow 1/T(\pi_t, t)$ \tcp*{Service rate} \label{line:drift_start}
       $\Delta(Q(t)) \leftarrow Q(t) \cdot (\lambda(t) - \mu_t)$ \; \label{line:drift_end}
       \tcp{Calculate shaped reward}
       $r_t \leftarrow -[V(t) \cdot \Delta(Q(t)) + w_T c_t^T + w_E c_t^E + w_A c_t^A]$ \; \label{line:reward}
       \tcp{Update queue dynamics}
       $Q(t+1) \leftarrow \max[Q(t) - \mu_t \cdot \delta_t, 0] + A(t)$ \; \label{line:queue}
       $\tau_k \leftarrow \tau_k \cup \{(s_t, a_t, r_t, s_{t+1})\}$\;
       $s_t \leftarrow s_{t+1}$\;
   }
   \textsc{StoreTrajectory}$(\mathcal{D}, \tau_k)$ \tcp*{Add to replay buffer}
   \If{$|\mathcal{D}| \geq B_{min}$ \textbf{and} $k \mod U_{freq} = 0$}{ \label{line:update_check}
       \tcp{Update critics}
       $\mathcal{B} \leftarrow$ \textsc{SampleBatch}$(\mathcal{D}, B_{size})$ \; \label{line:critic_start}
       $\mathcal{L}^{perf} \leftarrow \frac{1}{|\mathcal{B}|}\sum_{(s,r) \in \mathcal{B}} (V_\phi^{perf}(s) - R^{perf})^2$\;
       $\mathcal{L}^{stab} \leftarrow \frac{1}{|\mathcal{B}|}\sum_{(s,Q) \in \mathcal{B}} (V_\psi^{stab}(s) - Q_{target})^2$\;
       $\phi \leftarrow \phi - \alpha_\phi \nabla_\phi \mathcal{L}^{perf}$\;
       $\psi \leftarrow \psi - \alpha_\psi \nabla_\psi \mathcal{L}^{stab}$ \; \label{line:critic_end}
       \tcp{Update policy using PPO}
       \For{PPO epoch $e = 1, ..., E_{ppo}$}{ \label{line:ppo_start}
           $A^{total} \leftarrow$ \textsc{ComputeAdvantage}$(\mathcal{B}, V_\phi^{perf}, V_\psi^{stab})$\;
           $\rho \leftarrow \pi_\theta(a|s) / \pi_{old}(a|s)$ \tcp*{Importance ratio}
           $J_{PPO} \leftarrow \min(\rho A^{total}, \text{clip}(\rho, 1-\epsilon, 1+\epsilon) A^{total})$\;
           $\theta \leftarrow \theta + \alpha_\theta \nabla_\theta J_{PPO}$ \; \label{line:ppo_end}
       }
       \tcp{Adapt control parameters}
       $V(t) \leftarrow$ \textsc{UpdateControlParam}$(Q_{avg}, V)$ \; \label{line:adapt}
       $\tau_{temp} \leftarrow \beta \cdot \tau_{temp}$ \tcp*{Anneal exploration}
   }
}
\Return $\pi_\theta$
\end{algorithm}
\textit{Lines \ref{line:episode_start}-\ref{line:step_start} (Trajectory collection):} Each episode simulates a sequence of inference requests under varying network conditions. The policy $\pi_\theta$ generates partitioning decisions based on current system state, including queue backlog, network bandwidth, and resource availability.
\textit{Line \ref{line:execute} (Partition execution):} The selected partition $a_t$ is executed across edge and cloud, returning the next state and immediate costs $c_t = (c_t^T, c_t^E, c_t^A)$ for latency, energy, and accuracy.
\textit{Lines \ref{line:drift_start}-\ref{line:drift_end} (Lyapunov drift computation):} The drift captures the expected change in queue backlog. A positive drift indicates growing queues (instability), while negative drift indicates draining queues. This is the key innovation that ensures stability.
\textit{Line \ref{line:reward} (Reward shaping):} The reward combines immediate performance costs with the Lyapunov drift penalty. The control parameter $V(t)$ dynamically balances performance optimization against stability based on current queue state.
\textit{Line \ref{line:queue} (Queue evolution):} The queue dynamics follow the Lindley recursion, where $\delta_t$ is the time slot duration and $A(t)$ represents new arrivals following a Poisson process.
\textit{Lines \ref{line:critic_start}-\ref{line:critic_end} (Critic updates):} Two separate critics learn to predict performance costs and queue evolution. The performance critic estimates cumulative latency/energy costs, while the stability critic predicts future queue backlogs. Both methods utilize temporal difference learning with target networks to enhance stability.
\textit{Lines \ref{line:ppo_start}-\ref{line:ppo_end} (PPO policy update):} The policy is updated using Proximal Policy Optimization with a clipped surrogate objective to prevent destructive updates. The advantage function combines both critics' predictions weighted by $V(t)$.

\textit{Line \ref{line:adapt} (Control parameter adaptation):} The control parameter $V$ is adjusted based on average queue length to maintain stability while maximizing performance.

\subsection{Convergence and Stability Analysis}
\textbf{Theoretical guarantees.} Under our framework, we establish two key theoretical results:

\textit{Theorem 1 (Queue Stability):} If the arrival rate $\lambda < \mu_{max}$ where $\mu_{max}$ is the maximum achievable service rate, then the Lyapunov-assisted RL policy ensures:
\begin{equation}
\limsup_{T \to \infty} \frac{1}{T}\sum_{t=0}^{T-1} \mathbb{E}[Q(t)] \leq \frac{B + V \cdot g^*}{\epsilon}
\end{equation}
where $g^*$ is the optimal performance cost and $\epsilon = \mu_{max} - \lambda$ is the capacity margin.

\textit{Proof sketch:} By adding Lyapunov drift into the reward, the policy learns to take actions that minimize drift when queues grow large, ensuring bounded time-average backlog.

\textit{Theorem 2 (Performance Bound):} The time-average performance cost under our policy satisfies:
\begin{equation}
\limsup_{T \to \infty} \frac{1}{T}\sum_{t=0}^{T-1} \mathbb{E}[g(s_t, a_t)] \leq g^* + \frac{B}{V}
\end{equation}

This shows that performance approaches optimal as $V \to \infty$, with a tradeoff against queue backlog.

\subsection{Online Adaptation}

\textbf{Fast adaptation mechanism.} During deployment, network conditions and workloads may differ from training. We implement online adaptation through:
\begin{equation}
\theta_{adapted} = \theta_{base} + \alpha_{adapt} \nabla_\theta J_{online}
\end{equation}

where $J_{online}$ is computed from recent deployment experience with a higher weight on stability to prevent system degradation during adaptation.

\section{Implementation}

\subsection{System Architecture}\label{ssub: System architecture}

\begin{figure}
    \centering
    \includegraphics[width=1\linewidth]{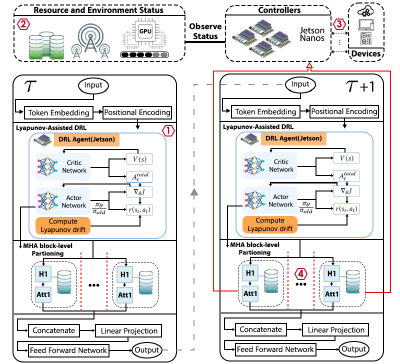}
    \caption{System architecture of \texttt{Splitwise}.
    The framework consists of four main runtime components (see Section \ref{ssub: System architecture}):
    (1) Partition Controller,
    (2) Profiling Engine,
    (3) Communication Manager,
    and (4) Execution Runtime.}
    \label{fig:Splitwise-arch}
\end{figure}

\textbf{Runtime components.} \texttt{Splitwise} consists of four main runtime components deployed across edge and cloud as shown in Figure \ref{fig:Splitwise-arch}: (1) \textit{Partition Controller} on the edge device that executes the learned policy and coordinates execution, (2) \textit{Profiling Engine} that collects performance metrics with minimal overhead, (3) \textit{Communication Manager} that handles data serialization and transmission with adaptive compression, and (4) \textit{Execution Runtime} on both edge and cloud that manages model shard execution. The controller maintains a lightweight state machine to track partition decisions and synchronize edge-cloud execution.

\textbf{Model preparation.} To enable fine-grained partitioning, we modify the model architecture at deployment time without retraining. Each transformer layer is decomposed into independently executable components \cite{10.1145/3037697.3037698}:
\begin{equation}
\text{Layer}_\ell = \{\text{Head}_1^{(\ell)}, ..., \text{Head}_H^{(\ell)}, \text{FFN}_1^{(\ell)}, \text{FFN}_2^{(\ell)}\}
\end{equation}
We implement custom CUDA kernels that allow individual attention heads to execute independently while maintaining numerical equivalence to the original model. The decomposition adds negligible overhead (<0.3\%) compared to monolithic execution.

We use an experience buffer of the 1000 most recent measurements \cite{10994299} and update the predictor every 100 inferences to adapt to changing system conditions.

\textbf{Asynchronous pipeline execution.} We implement a three-stage pipeline to hide communication latency \cite{butler2024pipeinfer}:
1) Edge computes partition $\pi_{edge}^{(\ell)}$ for layer $\ell$ then
2) While transmitting results to the cloud, the edge begins layer $\ell+1$ computation if $\pi_{edge}^{(\ell+1)} \neq \emptyset$ then
3) Cloud processes received data in parallel with edge execution.
This pipeline reduces effective latency by up to 35\% for balanced partitions where both the edge and cloud have substantial workloads.

\textbf{Dynamic model loading.} Edge devices cannot hold entire models in memory. We implement a dynamic loading scheme that maintains only active partitions:
\begin{equation}
M_{active} = \sum_{\ell=1}^{L} \sum_{h \in \mathcal{E}_h^{(\ell)}} m_h^{(\ell)} + \sum_{\ell: \pi_{FFN}^{(\ell)} \in \{0,2\}} m_{FFN}^{(\ell)}
\end{equation}
Model shards are loaded from flash storage with prefetching based on predicted future partitions, achieving \textless5ms loading latency for individual components.

\textbf{Partition checkpointing.} We maintain checkpoints at partition boundaries to enable recovery from communication failures:
\begin{equation}
\text{Checkpoint}_b = \{X_b, \pi_b, t_b\}
\end{equation}

If transmission fails, execution resumes from the last checkpoint with exponential backoff.

\section{Performance Evaluation} \label{sec:performance}
\subsection{Experimental Setup}

\textbf{Hardware platforms.} We evaluate \texttt{Splitwise} across diverse edge devices representing different deployment scenarios, from mobile phones to IoT gateways. Table~\ref{tab:hardware} summarizes our hardware configurations. The edge devices span a wide range of computational capabilities from the powerful Jetson Orin NX, designed for AI workloads, to the resource-constrained Raspberry Pi 5, which represents IoT scenarios. For cloud infrastructure, we utilize a university cluster, which is similar to an AWS EC2 p4d.24xlarge instance, equipped with 8 NVIDIA A100 GPUs, providing 640GB of GPU memory and 2.4TB/s of memory bandwidth. This setup reflects realistic edge-cloud deployments where resource-constrained devices collaborate with powerful cloud servers.

\begin{table}[b]
\centering
\caption{Hardware platforms used in evaluation}
\label{tab:hardware}
\resizebox{\linewidth}{!}{%
\begin{tabular}{llcccc} 
\hline
\rowcolor[rgb]{0.816,0.816,0.816} \textbf{Type} & \textbf{Device} & \textbf{Memory} & \textbf{Compute} & \textbf{Power} & \textbf{Use Case} \\ 
\hline
\multirow{3}{*}{Edge} & NVIDIA Jetson Orin NX & 8GB & 100 TOPS & 25W & AI Gateway \\
 & Samsung Galaxy S23 & 12GB & Snapdragon 8G2 & 8W & Mobile \\
 & Raspberry Pi 5 + NPU & 8GB & 13 TOPS & 12W & IoT Device \\ 
\hline
Cloud & AWS p4d.24xlarge & 1.1TB & 8$\times$A100 GPUs & - & Server \\
\hline
\end{tabular}
}
\end{table}

\textbf{Network conditions.} We emulate realistic network environments using Linux traffic control (\texttt{tc}) to shape bandwidth and latency between edge and cloud devices. Table~\ref{tab:network} presents our network configurations, derived from real-world 5G and WiFi measurements collected over 3 days in the university. The configurations capture typical scenarios from excellent WiFi connectivity to degraded cellular conditions.

\begin{table}[h]
\centering\scriptsize
\caption{Network conditions for evaluation. }
\label{tab:network}
\begin{tabular}{lcccc} 
\hline
\rowcolor[rgb]{0.816,0.816,0.816} \textbf{Scenario} & \textbf{Bandwidth} & \textbf{Latency} & \textbf{Jitter} & \textbf{Loss Rate} \\ 
\hline
Wifi & 100 Mbps & 10ms & 2ms & 0.01\% \\
5G-(Good) & 50 Mbps & 20ms & 5ms & 0.1\% \\
5G-(Average) & 25 Mbps & 40ms & 10ms & 0.5\% \\
4G & 10 Mbps & 80ms & 20ms & 1\% \\
Variable (Var) \cite{konrad2001markov}& 10-100 Mbps & 10-100ms & 2-20ms & 0.01-1\% \\
\hline
\end{tabular}

\end{table}
\textbf{Models and datasets.} Table~\ref{tab:models} details the LLM architectures used in our evaluation. We select models spanning three orders of magnitude in size to demonstrate \texttt{Splitwise}'s scalability. Each model presents unique challenges: GPT-2 1.5B \cite{radford2019language} fits entirely in edge memory but requires optimization for latency, LLaMA-7B\footnote{\scriptsize\url{https://huggingface.co/dfurman/LLaMA-7B}} necessitates careful memory management and partitioning, while LLaMA-13B\footnote{\scriptsize\url{https://huggingface.co/dfurman/LLaMA-13B}} cannot run on edge without our partitioning approach. The varying headcounts (16-40) and layer depths (24-40) test our framework's ability to handle diverse architectural patterns. We utilize the LMSYS-Chat-1M dataset~\cite{zheng2023lmsys}, which comprises one million real-world conversations from the Vicuna demo and ChatGPT interactions. This is a realistic inference workload with sequence lengths ranging from 50 to 2048 tokens.

\begin{table}[t]
\centering
\caption{Model configurations and characteristics}
\label{tab:models}
\resizebox{\linewidth}{!}{%
\begin{tabular}{lccccc} 
\hline
\rowcolor[rgb]{0.816,0.816,0.816} \textbf{Model} & \textbf{Parameters} & \textbf{Layers} & \textbf{Heads} & \textbf{Hidden} & \textbf{Memory} \\ 
\hline
GPT-2 \cite{radford2019language} & 1.5B & 24 & 16 & 1,600 & 6GB \\
LLaMA-7B & 7B & 32 & 32 & 4,096 & 28GB \\
LLaMA-13B & 13B & 40 & 40 & 5,120 & 52GB \\
\hline
\end{tabular}
}
\end{table}

\textbf{Baselines.}  To ensure fair comparison, we implement all baselines using their optimal reported configurations.  
All methods share identical hardware platforms and network conditions during evaluation. We use the authors' official implementations where available or reproduce following published specifications. All experiments use identical seeds. Table~\ref{tab:baseline_config} details the specific settings for \texttt{Splitwise}.

\textit{Edge-only}: Entire model on edge with 4-bit quantization,
\textit{Cloud-only}: Full model on cloud with network transmission,
\textit{Edgeshard}: Fixed partition at layer $L/2$ \cite{zhang2024edgeshard},
\textit{PipeEdge}: Pipeline parallelism with static optimization \cite{9996638},
\textit{CE-LSLM}: Dynamic execution with early exit \cite{zhu2025lslm}.

\begin{table}
\centering
\scriptsize
\setlength{\extrarowheight}{0pt}
\setlength{\aboverulesep}{0pt}
\setlength{\belowrulesep}{0pt}
\caption{\texttt{Splitwise} hyperparameter configurations and training settings}
\label{tab:baseline_config}
\resizebox{\linewidth}{!}{%
\begin{tabular}{clc} 
\toprule
\rowcolor[rgb]{0.816,0.816,0.816} \textbf{Category} & \textbf{Hyperparameter} & \textbf{Value} \\ 
\midrule
\multirow{6}{*}{\textbf{RL Training}} & Learning rate (actor) & $3 \times 10^{-4}$ \\
 & Learning rate (critics) & $1 \times 10^{-3}$ \\
 & Discount factor $\gamma$ & 0.99 \\
 & GAE parameter $\lambda$ & 0.95 \\
 & PPO clip range & 0.2 \\
 & Entropy coefficient & 0.01 \\
\midrule
\multirow{5}{*}{\textbf{Lyapunov Control}} & V range & [0.1, 10.0] \\
 & V adaptation rate & 0.1 \\
 & Reference queue $Q_{ref}$ & 10 requests \\
 & Critical queue $Q_{critical}$ & 50 requests \\
 & Drift bound $B$ & 100 \\
\midrule
\multirow{5}{*}{\textbf{Experience Replay}} & Replay buffer size & 1,000 transitions \\
 & Batch size & 256 \\
 & Priority exponent $\alpha$ & 0.6 \\
 & Update frequency & Every 100 steps \\
\midrule
\multirow{4}{*}{\textbf{Training Process}} & Episodes & 2,500 \\
 & Warm-up episodes & 100 \\
 & Evaluation frequency & Every 50 episodes \\
\bottomrule
\end{tabular}
}
\end{table}

\textbf{Metrics.} We measure: (i) end-to-end latency (P50, P95, P99), (ii) energy consumption on edge device, (iii) model accuracy on dataset, and (iv) partitioning in various networks.

\subsection{Experimental Results}

\textbf{Ablation Study and Scaling Analysis.}
The ablation study in Table~\ref{tab:ablation} and the scaling analysis in Table~\ref{tab:scaling} collectively demonstrate the effectiveness, robustness, and scalability of the \texttt{Splitwise} framework. Table~\ref{tab:ablation} evaluates the contribution of key components by measuring end-to-end latency and system stability under varying configurations. The full \texttt{Splitwise} system achieves a latency of 87 ms while maintaining stability at a request rate of 8.5 requests per second. The latency increases by 4.6\% when we remove the Lyapunov drift from the reward function. More critically, system instability at just 6 requests per second underscores the crucial role of Lyapunov optimization in ensuring queue stability under dynamic workloads. Disabling adaptive $V$ significantly degrades performance, increasing latency by 9\% to 95 ms, as the system loses its ability to minimize communication overhead at partition boundaries. 

Table~\ref{tab:scaling} illustrates how \texttt{Splitwise} scales with increasing model size, enabling efficient inference of large LLMs on edge devices. As the model grows from 1.5B to 13B parameters, \texttt{Splitwise} achieves increasing speedups (on average from 1.4$\times$ to 2.8$\times$) by offloading computationally intensive components to the cloud while keeping memory and communication demands within feasible limits for edge devices. The required edge memory increases from 2.1 GB to 5.8 GB, remaining below the capacity of modern edge hardware while communication volume scales sublinearly with model size.
As shown in Table~\ref{tab:scaling}, the accuracy degradation is small
(less than 4\% from GPT-2 1.5B to LLaMA-13B).
This decrease mainly comes from two factors:
(i) lightweight activation quantization used only at edge-cloud boundaries to lower transmission costs, and 
(ii) minor numerical variance introduced during distributed recomposition of multi-head attention outputs.

\begin{table}
\centering\scriptsize
\setlength{\extrarowheight}{0pt}
\setlength{\aboverulesep}{0pt}
\setlength{\belowrulesep}{0pt}
\caption{Ablation Study: component contributions}
\label{tab:ablation}
\begin{tabular}{lcc} 
\toprule
\rowcolor[rgb]{0.816,0.816,0.816} \textbf{Configuration} & \textbf{Latency (ms)} & \textbf{Stability}  \\ 
\midrule
Full \texttt{Splitwise} & 87 & Stable at 8.5 req/s \\
w/o Lyapunov drift & 91 & Unstable at 6 req/s \\
w/o Pipeline exec. & 118 & Stable at 8.5 req/s \\
Fixed V parameter & 95 & Stable at 7.2 req/s \\
\bottomrule
\end{tabular}
\end{table}

\begin{table}[b]
\centering\scriptsize
\setlength{\extrarowheight}{0pt}
\setlength{\aboverulesep}{0pt}
\setlength{\belowrulesep}{0pt}
\caption{Performance scaling with model size}
\label{tab:scaling}
\begin{tabular}{lcccc} 
\toprule
\rowcolor[rgb]{0.816,0.816,0.816} \textbf{Model Size} & \textbf{Speedup} & \textbf{Memory (GB)} & \textbf{Comm. (GB)} & \textbf{Accuracy} \\ 
\midrule
1.5B & 1.4$\times$ & 2.1 & 0.8 & 91.2\% \\
7B & 2.3$\times$ & 4.2 & 2.1 & 90.5\% \\
13B & 2.8$\times$ & 5.8 & 3.5 & 87.8\% \\
\bottomrule
\end{tabular}
\end{table}

\begin{figure}[!]
    \centering
    \begin{subfigure}[t]{0.49\columnwidth}
        \centering
            \includegraphics[width=1\linewidth]{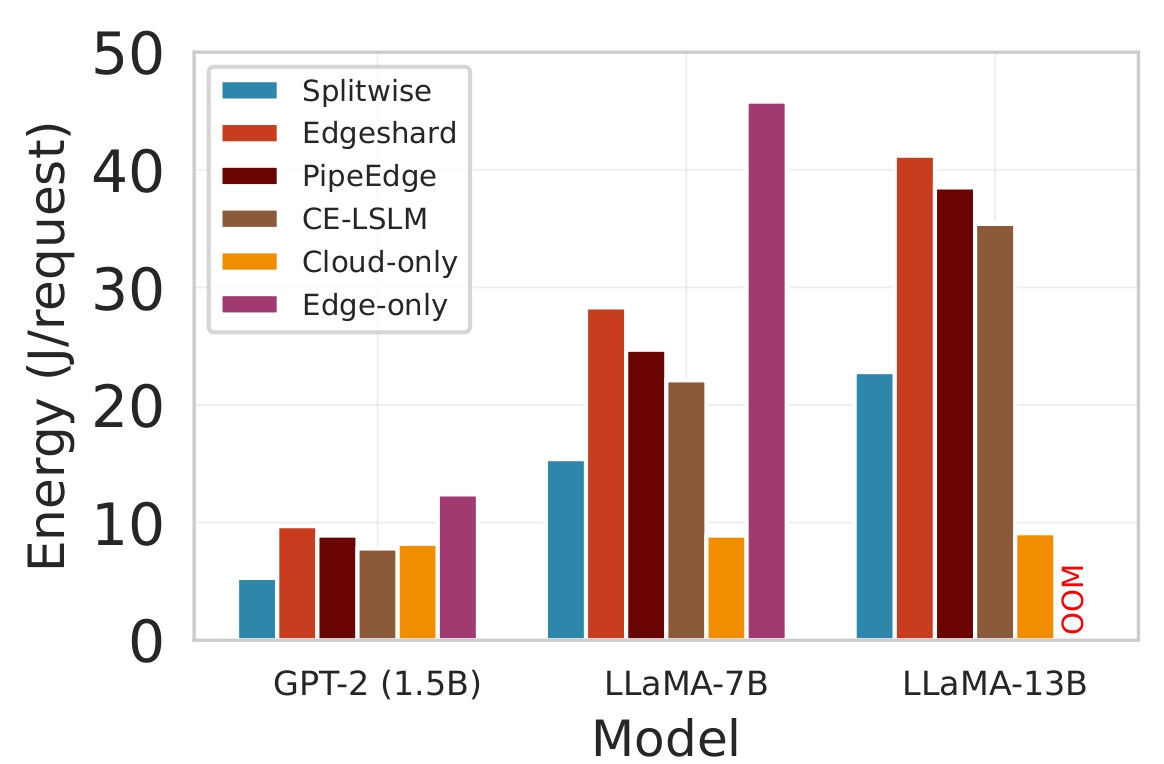}
        \vspace{-0.6cm}
        \caption{Jetson Orin energy}
        \label{fig:ms-cyber}
    \end{subfigure}
    \begin{subfigure}[t]{0.49\columnwidth}
        \centering
            \includegraphics[width=1\linewidth]{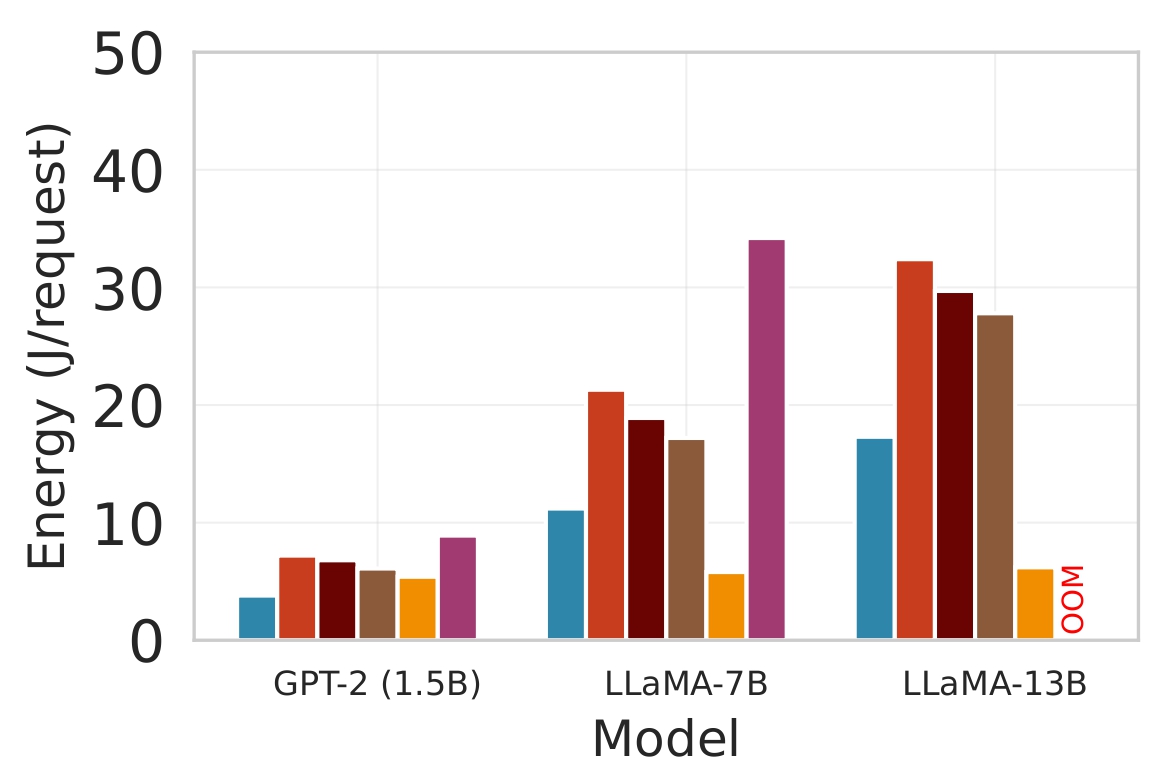}
        \vspace{-0.6cm}
        \caption{Galaxy s23 energy}
        \label{fig:ms-epi}
    \end{subfigure}

    \begin{subfigure}[t]{0.49\columnwidth}
        \centering
            \includegraphics[width=1\linewidth]{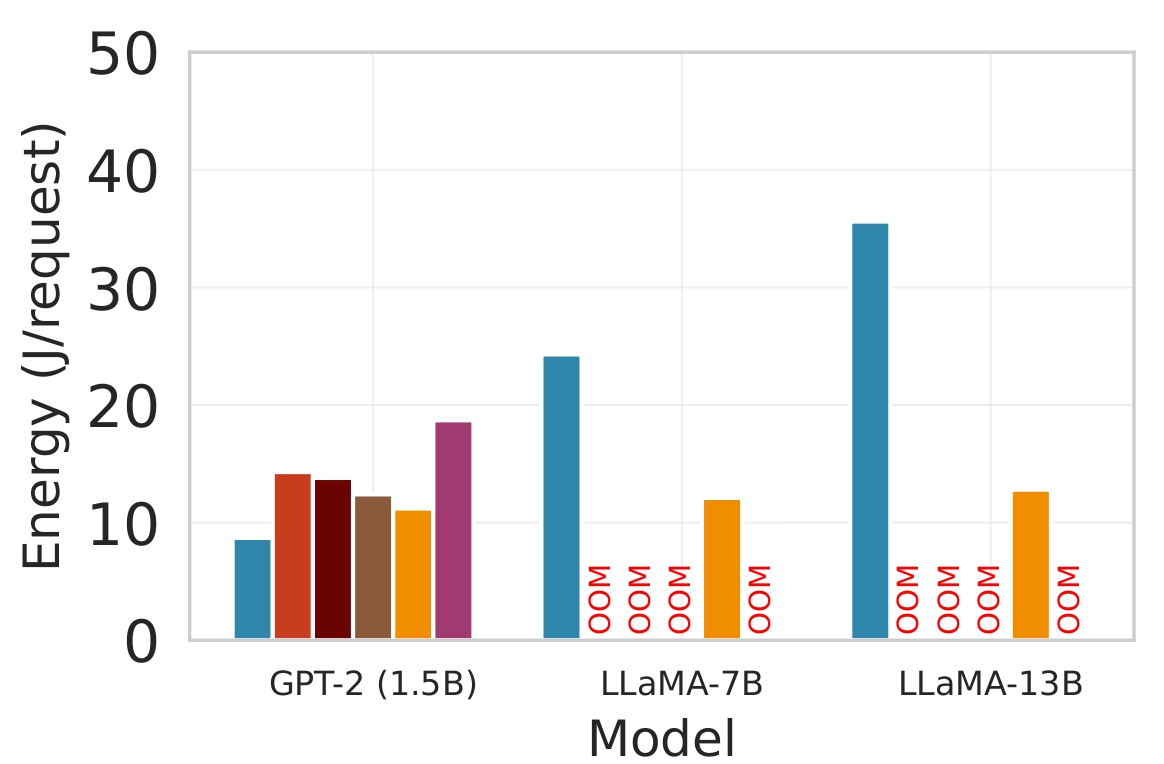}
        \vspace{-0.6cm}
        \caption{Raspberry energy}
        \label{fig:ms-inspiral}
    \end{subfigure}
\caption{Energy consumption comparison across different edge devices and LLM models on variable network conditions. Energy per request (a) on the NVIDIA Jetson Orin NX, (b) on the Samsung Galaxy S23 smartphone, and (c) on the Raspberry Pi 5.}

    \label{fig:energy}
\end{figure}

\begin{figure}[t]
        \centering
            \includegraphics[width=0.6\linewidth]{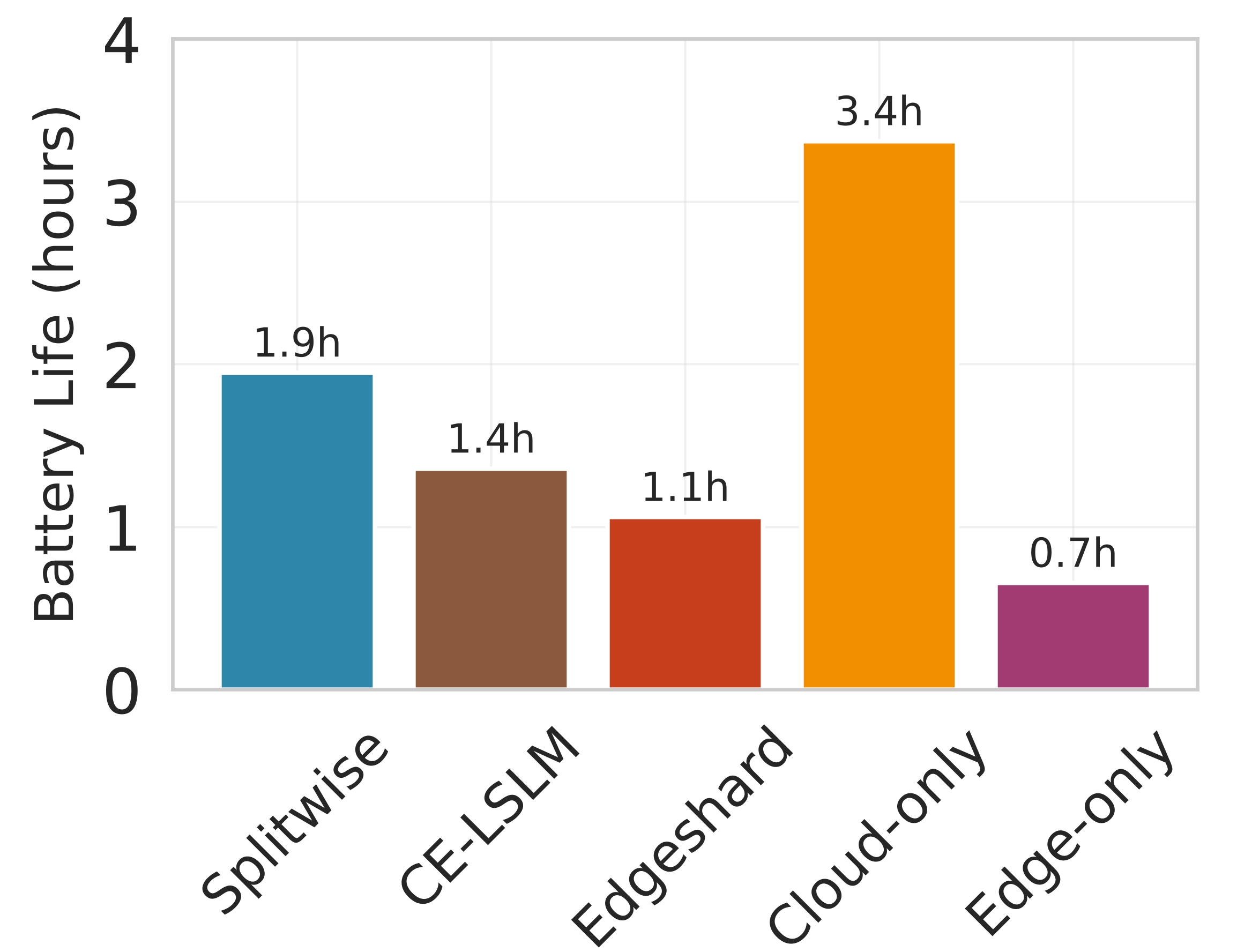}
        \caption{Mobile device battery life with 600 inference requests/hour.}
        \label{fig:battery}
\end{figure}

\textbf{Energy Consumption.} 
Figure~\ref{fig:energy} presents a comprehensive analysis of the energy efficiency of \texttt{Splitwise} compared to state-of-the-art baselines across three diverse edge devices with varying computational capabilities: the powerful NVIDIA Jetson Orin NX (designed for AI workloads), the mobile Samsung Galaxy S23, and the resource-constrained Raspberry Pi 5. The evaluation is conducted using three LLMs of increasing size to assess scalability.

The results consistently show that \texttt{Splitwise} achieves lower energy consumption than all competing approaches. On the Jetson Orin (Figure~\ref{fig:ms-cyber}), \texttt{Splitwise} reduces energy by up to 41\% compared to Edgeshard and PipeEdge, and by over 77\% compared to the Edge-only baseline for the LLaMA-7B model. This substantial improvement is attributed to its intelligent partitioning strategy, which offloads computationally intensive components to the cloud, thereby reducing the load on the edge device. The Cloud-only approach exhibits low energy consumption due to minimal local computation. It is impractical for sensitive applications due to privacy concerns and high latency. The Edge-only baseline, particularly for larger models like LLaMA-7B and LLaMA-13B, incurs very high energy costs because it must execute the entire model locally without using the cloud's superior computational resources.
On the Galaxy S23 (Figure~\ref{fig:ms-epi}), the trend is similar, with \texttt{Splitwise} consuming less energy than all other methods. Notably, the energy savings are even more pronounced for the larger models, highlighting the benefits of collaborative inference for power-efficient execution on mobile devices. The Raspberry Pi 5 (Figure~\ref{fig:ms-inspiral}) represents the most challenging scenario due to its limited memory and processing power. The Edge-only approach fails to run LLaMA-7B and LLaMA-13B entirely, as indicated by "OOM" (Out of Memory) errors. While \texttt{Splitwise} can successfully run these large models, it does so with significantly higher energy consumption compared to smaller models. This is because the dynamic loading and partitioning process introduces additional overhead, and the system must frequently swap model shards between flash storage and RAM.

Figure~\ref{fig:battery} demonstrates the impact of different inference strategies on mobile device battery longevity, a critical factor for user experience and practical deployment. Under a continuous workload of 600 requests per hour, the Cloud-only baseline achieves the longest battery life of 3.4 hours. This is because it offloads all computation to the cloud, minimizing local CPU and GPU usage on the edge device \cite{10.1007/978-3-030-29400-7_21}. However, this approach introduces significant latency and cost.

\begin{figure}[!]
    \centering
    \begin{subfigure}[t]{0.49\columnwidth}
        \centering
            \includegraphics[width=1\linewidth]{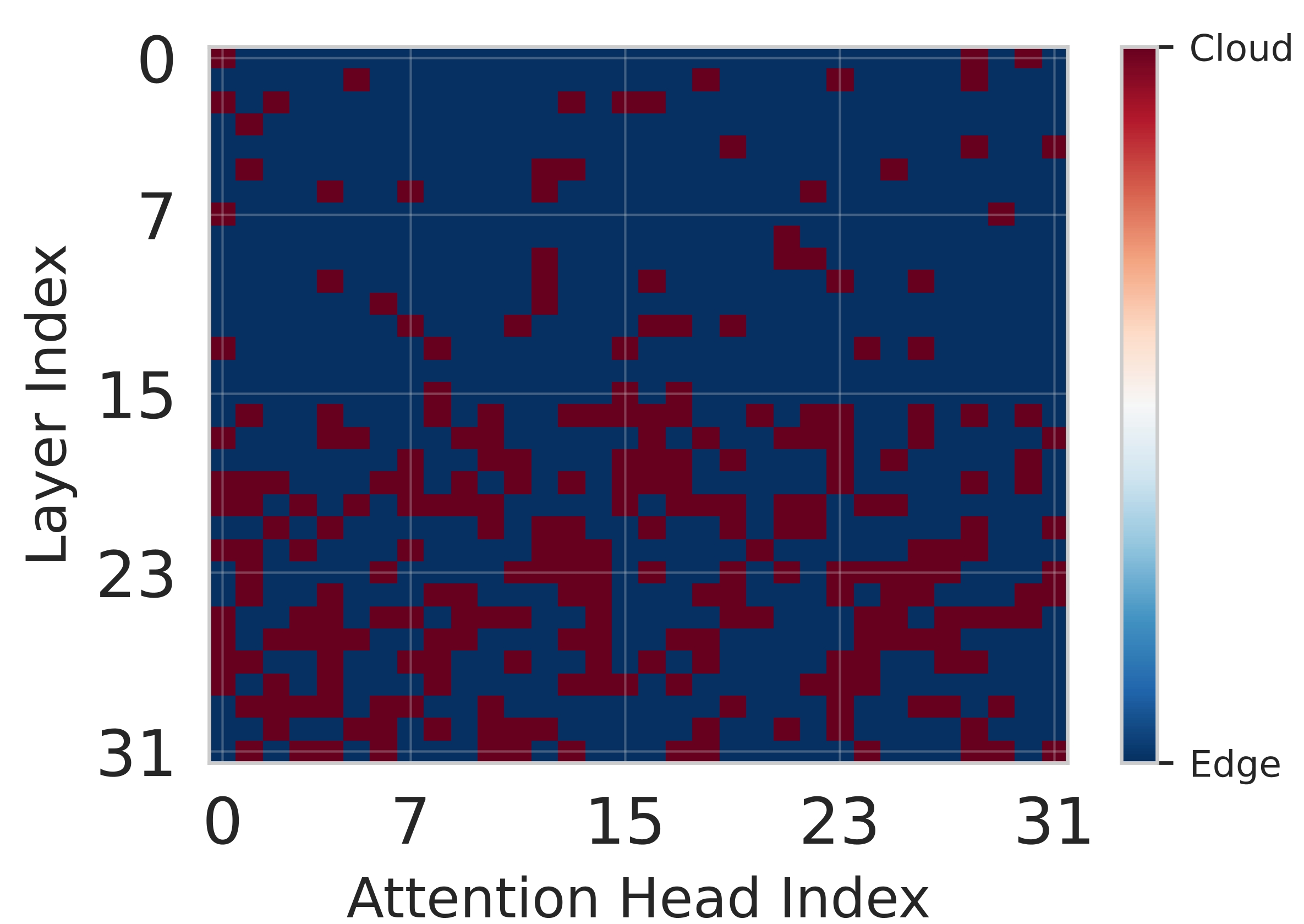}
        \vspace{-0.6cm}
        \caption{Poor Network (10 Mbps)}
        \label{fig:Poor Network (10 Mbps}
    \end{subfigure}
    \begin{subfigure}[t]{0.49\columnwidth}
        \centering
            \includegraphics[width=1\linewidth]{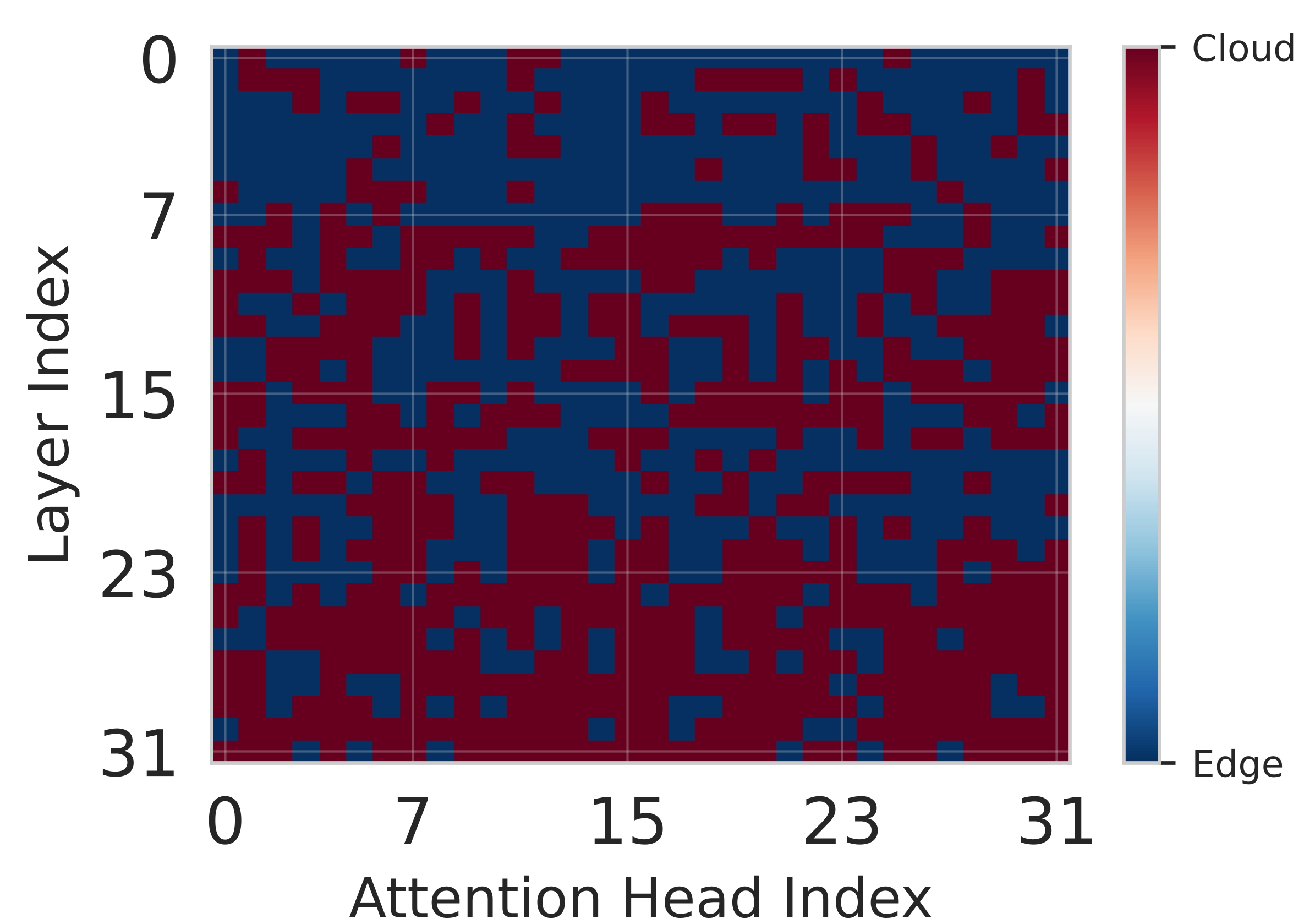}
        \vspace{-0.6cm}
        \caption{Good Network (100 Mbps)}
        \label{fig:Good Network (100 Mbps)}
    \end{subfigure}

    \begin{subfigure}[t]{0.49\columnwidth}
        \centering
            \includegraphics[width=1\linewidth]{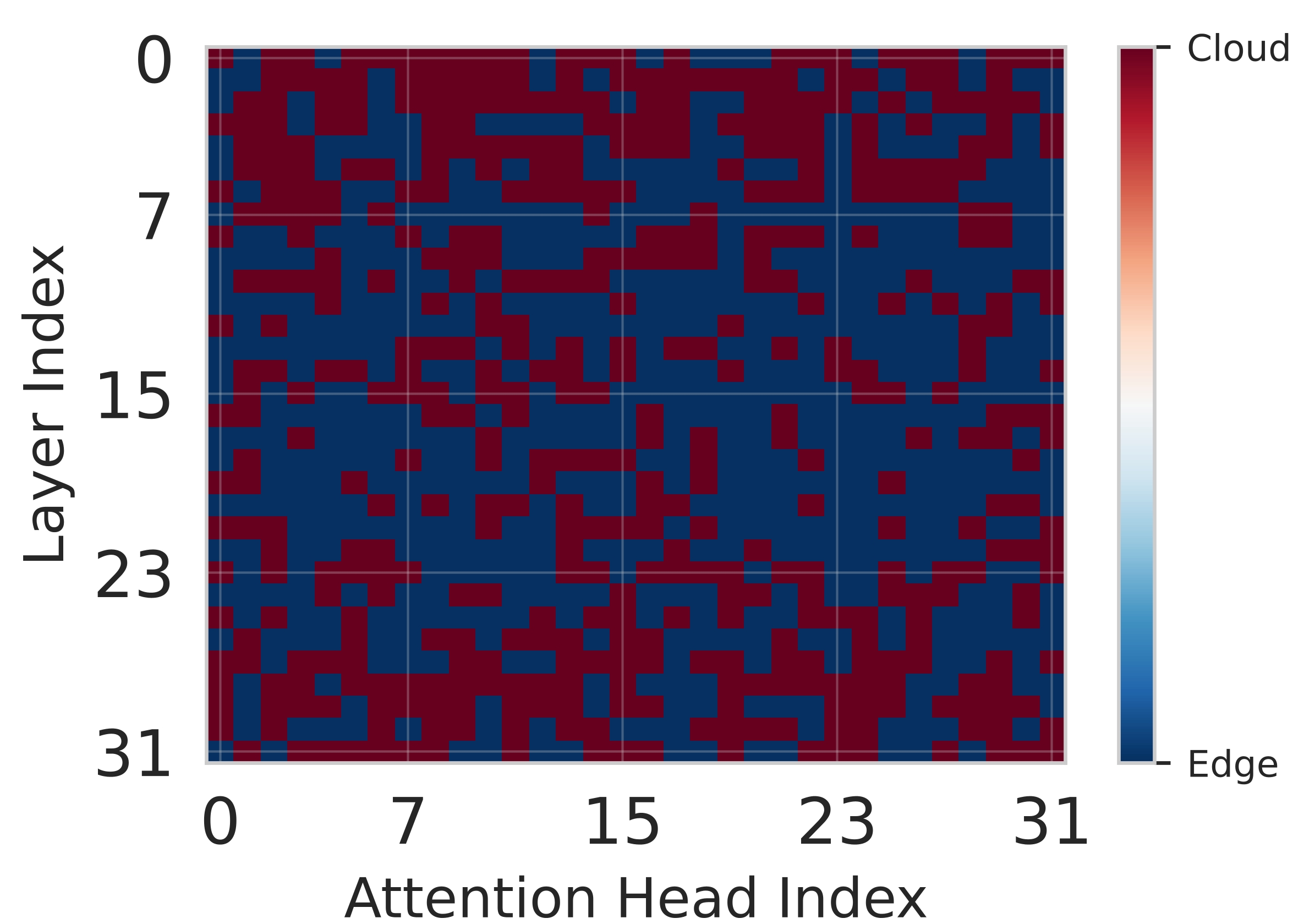}
        \vspace{-0.6cm}
        \caption{Variable Network}
        \label{fig:Variable Network}
    \end{subfigure}

     \caption{Dynamic and fine-grained partitioning strategies learned by \texttt{Splitwise} 
    under (a) poor, (b) good, and (c) variable network conditions. 
    Each heatmap illustrates the placement of transformer attention heads across layers, 
    where blue indicates edge execution and red denotes cloud execution.}
    \label{fig:network}
\end{figure}

\textbf{Network.} The heatmaps in Figure~\ref{fig:network} illustrate the fine-grained partitioning policies learned by \texttt{Splitwise} across different network scenarios, highlighting its dynamic adaptation capabilities. In the poor network condition with 10 Mbps bandwidth (Figure~\ref{fig:Poor Network (10 Mbps}), the policy exhibits a strong bias towards edge execution, as evidenced by the prevalence of blue pixels. This conservative strategy minimizes communication overhead, which is critical when network bandwidth is limited. By keeping computation local, \texttt{Splitwise} prioritizes low latency and reduces energy consumption associated with data transmission, effectively mitigating the significant delays that would otherwise be incurred by offloading work to the cloud. Conversely, under good network conditions with 100 Mbps bandwidth (Figure~\ref{fig:Good Network (100 Mbps)}), the partitioning becomes significantly more balanced and flexible. The complex interplay of red and blue regions indicates that the agent leverages the high-bandwidth link to offload computationally intensive components to the powerful cloud infrastructure while retaining less demanding computations on the edge. This approach optimizes overall system performance by exploiting the complementary strengths of both environments.

Finally, in the variable network scenario (Figure~\ref{fig:Variable Network}), which simulates real-world fluctuations in connectivity, the partitioning pattern reflects a cautious strategy that favors edge execution. This behavior is driven by the Lyapunov assisstance in \texttt{Splitwise}, which incorporates stability guarantees into the reward function. The agent learns to prioritize queue stability, ensuring bounded latency even during periods of poor network quality.

\begin{figure}
    \centering
    \begin{subfigure}[t]{0.49\columnwidth}
        \centering
            \includegraphics[width=1\linewidth]{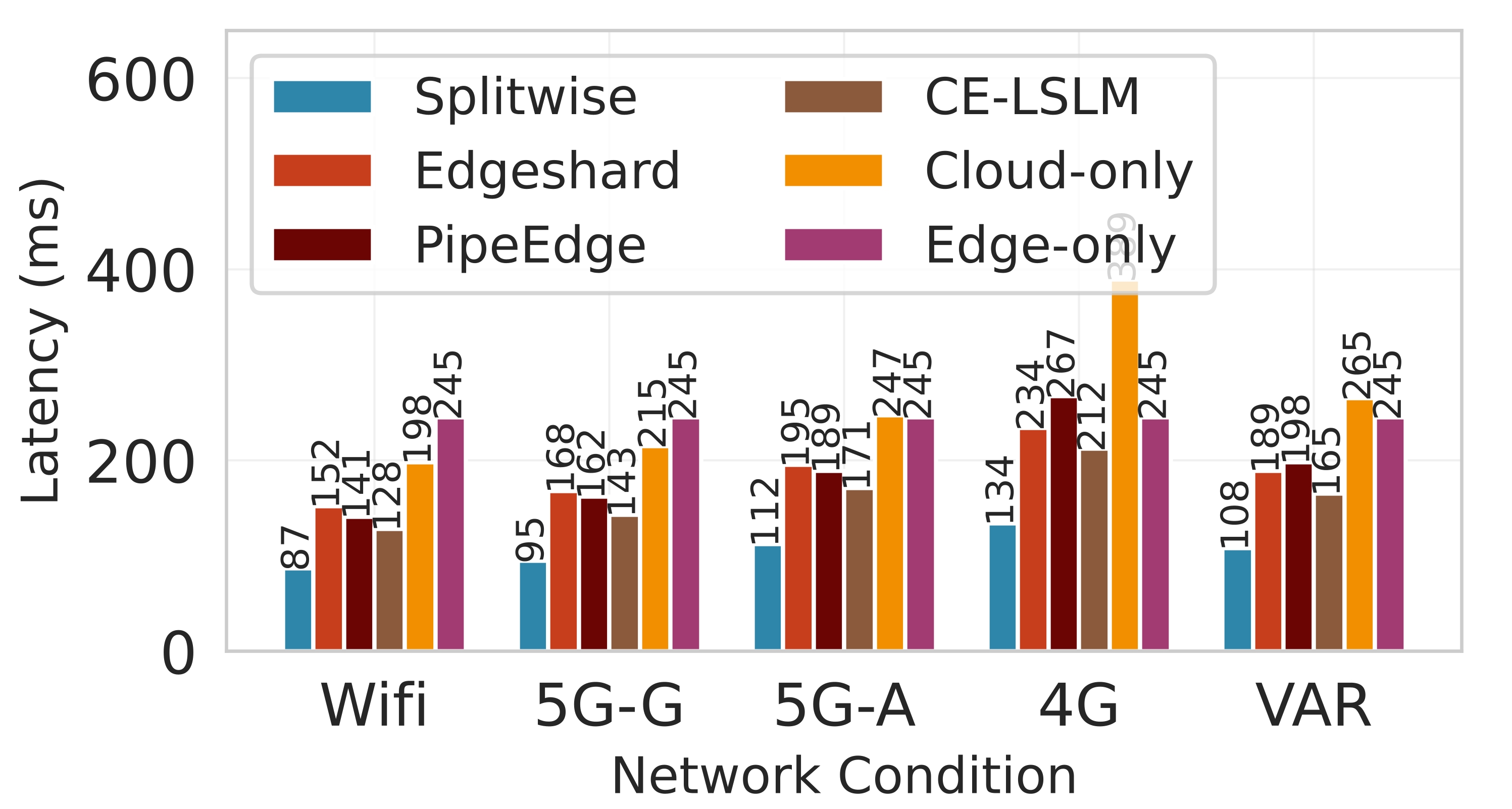}
        \vspace{-0.6cm}
        \caption{P50 Latency}
        \label{fig:P50 Latency}
    \end{subfigure}
    \begin{subfigure}[t]{0.49\columnwidth}
        \centering
            \includegraphics[width=1\linewidth]{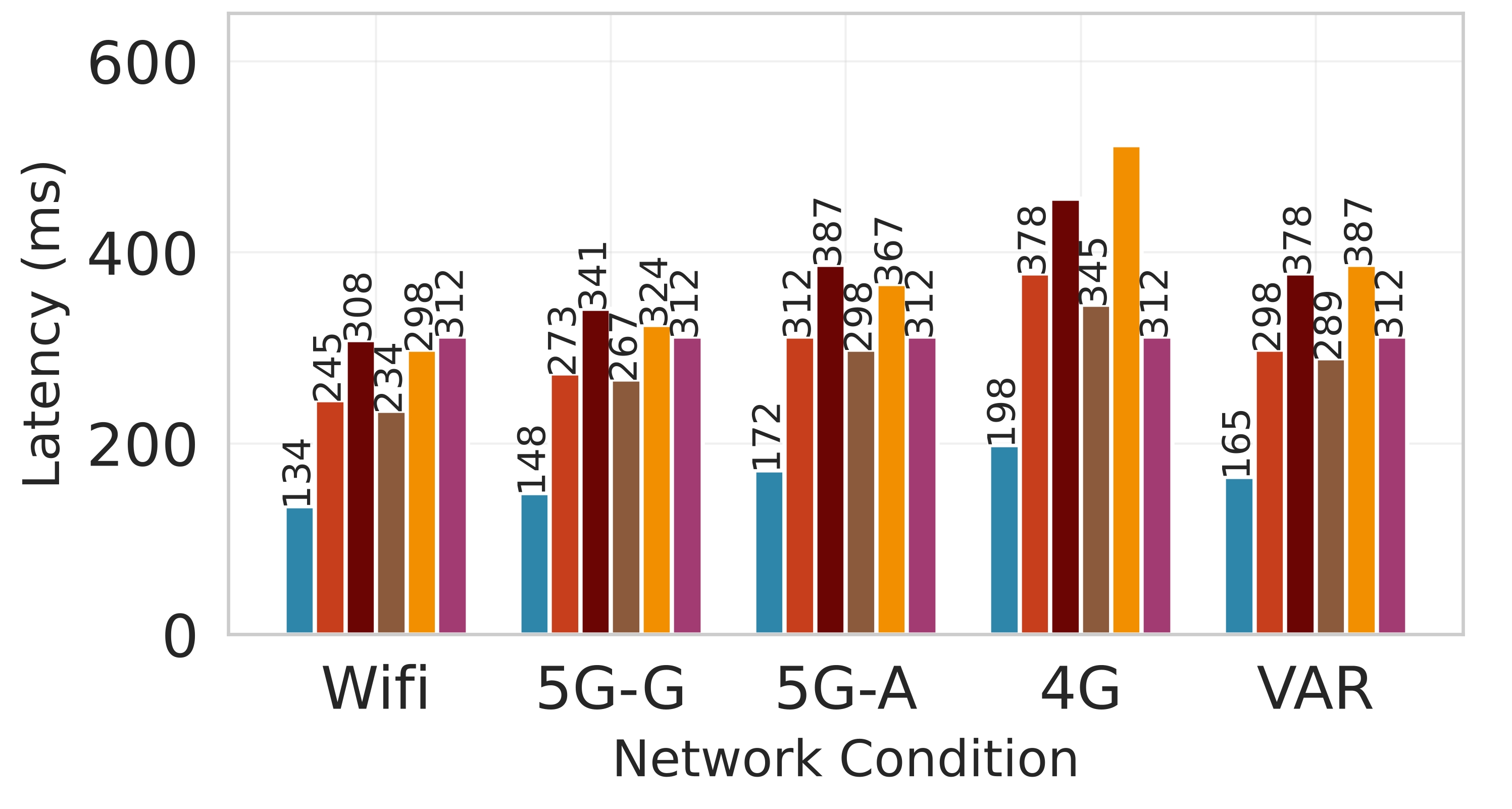}
        \vspace{-0.6cm}
        \caption{P95 Latency}
        \label{fig:P95 Latency}
    \end{subfigure}

    \begin{subfigure}[t]{0.49\columnwidth}
        \centering
            \includegraphics[width=1\linewidth]{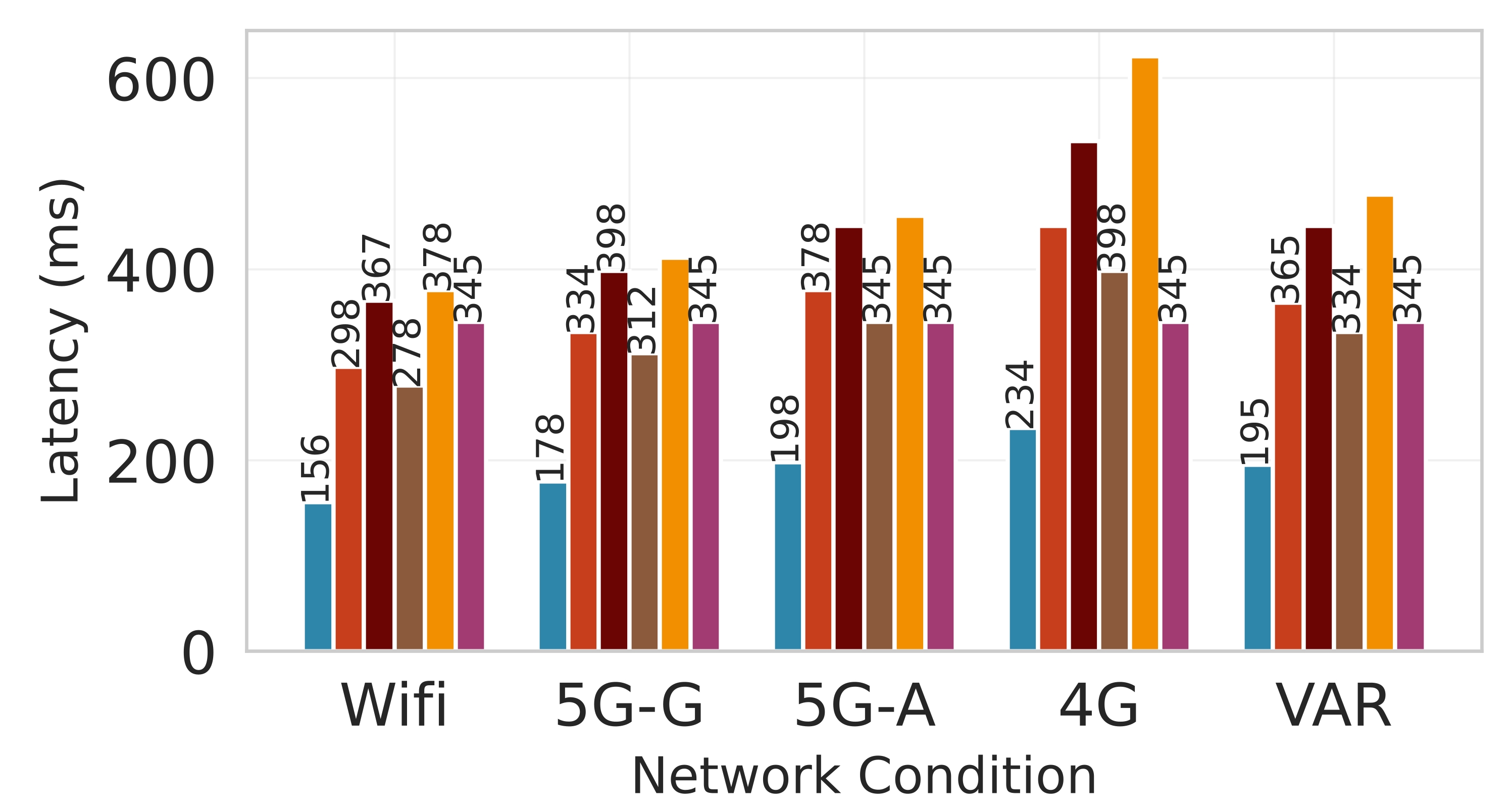}
        \vspace{-0.6cm}
        \caption{P99 Latency}
        \label{fig:P99 Latency}
    \end{subfigure}
    \begin{subfigure}[t]{0.49\columnwidth}
        \centering
            \includegraphics[width=1\linewidth]{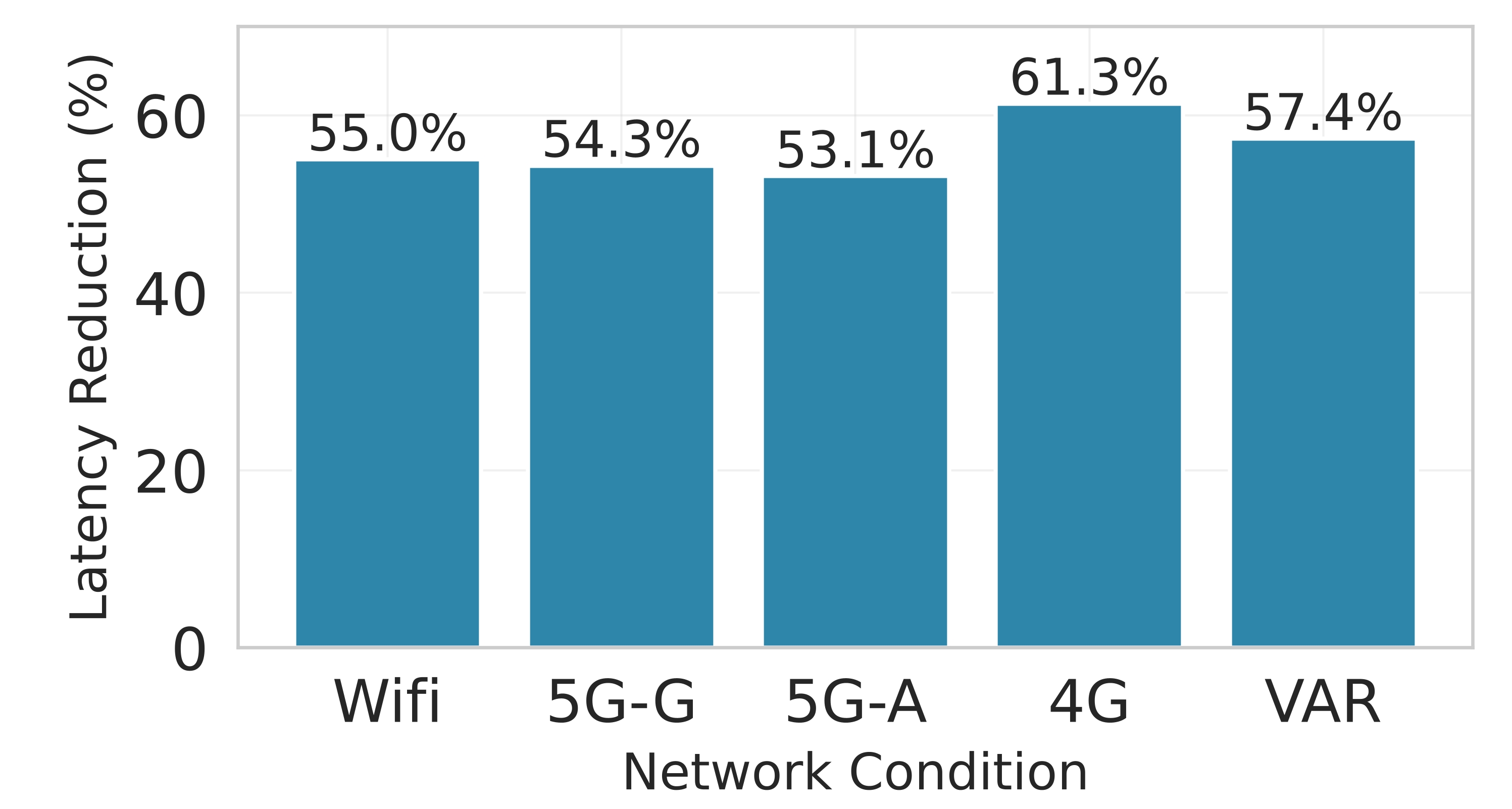}
        \vspace{-0.6cm}
        \caption{Latency Reduction vs Cloud-only}
        \label{fig:/Splitwise P95 Latency Reduction vs Cloud-only}
    \end{subfigure}
    \caption{Performance evaluation of \texttt{Splitwise} under diverse network conditions. (a) and (c) compare the median (P50) and 99th percentile (P99) end-to-end inference latency of \texttt{Splitwise} against state-of-the-art baselines across different network scenarios. (b) presents the P95 latency results. (d) quantifies the percentage reduction in P95 latency achieved by \texttt{Splitwise} compared to the Cloud-only baseline.}
    \label{fig:latency}
\end{figure}

\textbf{Latency.} Figure~\ref{fig:latency} shows the effectiveness of \texttt{Splitwise} in minimizing latency while maintaining system stability.
In both the P50 (median) and P99 (99th percentile) latency metrics shown in Figures~\ref{fig:P50 Latency} and~\ref{fig:P95 Latency}, \texttt{Splitwise} consistently outperforms all competing baselines across all network scenarios. The cloud-only approach, which incurs significant communication overhead, exhibits the highest latency, particularly under poor network conditions like 4G and VAR, where it can exceed 500 ms. In contrast, edge-only execution, while avoiding network delays, suffers from high computational latency due to the limited processing power of edge devices. Static partitioning methods such as Edgeshard and PipeEdge perform better than these extremes but still fail to adapt to dynamic network fluctuations, resulting in suboptimal performance.
The superiority of \texttt{Splitwise} is most shown in its ability to maintain low and stable latency even under challenging conditions. This is achieved through its dynamic, fine-grained partitioning strategy that employs the Lyapunov optimization framework to balance immediate performance with long-term queue stability. By intelligently allocating computation between the edge and cloud based on network quality, \texttt{Splitwise} minimizes the impact of network bottlenecks. Figure~\ref{fig:/Splitwise P95 Latency Reduction vs Cloud-only} demonstrates that \texttt{Splitwise} achieves a 53.1\% to 61.3\% reduction in P95 latency compared to the cloud-only baseline across all network types.

\section{Related Work} \label{sec:related_work}

\textbf{Static model partitioning.} Early work on DNN partitioning~\cite{10.1145/3037697.3037698,li2019edgeaiondemandaccelerating} pioneered layer-wise splitting between edge and cloud. Neurosurgeon~\cite{10.1145/3037697.3037698} profiles per-layer latency and energy costs offline and selects a single optimal split point. DADS \cite{8737614} extends this by considering multiple DNNs simultaneously but remains limited to static partitioning. These approaches fundamentally assume stable network conditions and uniform workload assumptions, which fail in real-world deployments where bandwidth varies the day~\cite{howard2017mobilenetsefficientconvolutionalneural}. Moreover, existing methods partition at layer granularity, missing opportunities for finer-grained optimization within transformer blocks.

\textbf{Edge-cloud collaborative inference.} Collaborative inference systems~\cite{malka2022decentralizedlowlatencycollaborativeinference,Laskaridis_2020,10.1145/3498361.3538932,chen2025adaptive,zhu2025lslm,he2024large,jin2024collm} distribute computation across edge and cloud resources. SPINN~\cite{Laskaridis_2020} progressively refines predictions using early exits, but this approach is incompatible with autoregressive LLMs, where each token depends on complete model execution. CoDL~\cite{10.1145/3498361.3538932} dynamically adjusts partition points for CNNs but relies on heuristic policies that fail to generalize across model architectures.

\textbf{\texttt{Splitwise} Features.} Our work differs from prior literature in 3 ways. First, we enable fine-grained partitioning at sub-layer granularity, exposing an order of magnitude more partition points. Second, we provide theoretical guarantees on queue stability through Lyapunov analysis while simultaneously optimizing multiple objectives. Third, we handle the combinatorial explosion of the action space through hierarchical decomposition and learned embeddings, making the approach tractable for large models.

\section{Conclusion} \label{sec:conclusion}
\texttt{Splitwise} demonstrates that fine-grained partitioning with queue-stability guarantees enables efficient edge–cloud LLM inference. It decomposes transformer layers into attention heads and feed-forward blocks and uses a hierarchical policy with Lyapunov-assisted rewards to assign components to either the edge or the cloud. This design exposes far more split options than layer‑wise methods while remaining tractable through action decomposition.
It adapts to fluctuating network links, delivering latency reductions of 1.4$\times$–2.8$\times$, up to 41\% energy savings, and 53–61\% lower P95 latency compared with static and cloud-only baselines. It balances immediate performance with long‑term queue stability.
Future work includes integrating early exits and compression to further cut communication and broaden applicability.

\begin{acks}
This paper is funded by the Recovery and Resilience Plan of the Slovak Republic, within the “Transformation and Innovation Consortia” call (project code: 09I02-03-V01-00012).
\end{acks}

\bibliographystyle{ACM-Reference-Format}
\bibliography{references}


\begin{thebibliography}{34}


\ifx \showCODEN    \undefined \def \showCODEN     #1{\unskip}     \fi
\ifx \showDOI      \undefined \def \showDOI       #1{#1}\fi
\ifx \showISBNx    \undefined \def \showISBNx     #1{\unskip}     \fi
\ifx \showISBNxiii \undefined \def \showISBNxiii  #1{\unskip}     \fi
\ifx \showISSN     \undefined \def \showISSN      #1{\unskip}     \fi
\ifx \showLCCN     \undefined \def \showLCCN      #1{\unskip}     \fi
\ifx \shownote     \undefined \def \shownote      #1{#1}          \fi
\ifx \showarticletitle \undefined \def \showarticletitle #1{#1}   \fi
\ifx \showURL      \undefined \def \showURL       {\relax}        \fi
\providecommand\bibfield[2]{#2}
\providecommand\bibinfo[2]{#2}
\providecommand\natexlab[1]{#1}
\providecommand\showeprint[2][]{arXiv:#2}

\bibitem[Bae et~al\mbox{.}(2020)]%
        {bae2020reinforcement}
\bibfield{author}{\bibinfo{person}{Sohee Bae}, \bibinfo{person}{Seungyul Han}, {and} \bibinfo{person}{Youngchul Sung}.} \bibinfo{year}{2020}\natexlab{}.
\newblock \showarticletitle{A reinforcement learning formulation of the Lyapunov optimization: Application to edge computing systems with queue stability}.
\newblock \bibinfo{journal}{\emph{arXiv preprint arXiv:2012.07279}} (\bibinfo{year}{2020}).
\newblock


\bibitem[Boateng et~al\mbox{.}(2025)]%
        {boateng2025survey}
\bibfield{author}{\bibinfo{person}{Gordon~Owusu Boateng} {et~al\mbox{.}}} \bibinfo{year}{2025}\natexlab{}.
\newblock \showarticletitle{A survey on large language models for communication, network, and service management: Application insights, challenges, and future directions}.
\newblock \bibinfo{journal}{\emph{IEEE Commun. Surv. Tutor.}} (\bibinfo{year}{2025}).
\newblock


\bibitem[Butler et~al\mbox{.}(2024)]%
        {butler2024pipeinfer}
\bibfield{author}{\bibinfo{person}{Branden Butler}, \bibinfo{person}{Sixing Yu}, \bibinfo{person}{Arya Mazaheri}, {and} \bibinfo{person}{Ali Jannesari}.} \bibinfo{year}{2024}\natexlab{}.
\newblock \showarticletitle{Pipeinfer: Accelerating llm inference using asynchronous pipelined speculation}. In \bibinfo{booktitle}{\emph{SC24: International Conference for High Performance Computing, Networking, Storage and Analysis}}. IEEE, \bibinfo{pages}{1--19}.
\newblock


\bibitem[Chapman et~al\mbox{.}(1994)]%
        {10.1007/3-540-57659-2_11}
\bibfield{author}{\bibinfo{person}{Barbara~M. Chapman}, \bibinfo{person}{Thomas Fahringer}, {and} \bibinfo{person}{Hans~P. Zima}.} \bibinfo{year}{1994}\natexlab{}.
\newblock \showarticletitle{Automatic support for data distribution on distributed memory multiprocessor systems}. In \bibinfo{booktitle}{\emph{Languages and Compilers for Parallel Computing}}, \bibfield{editor}{\bibinfo{person}{Utpal Banerjee}, \bibinfo{person}{David Gelernter}, \bibinfo{person}{Alex Nicolau}, {and} \bibinfo{person}{David Padua}} (Eds.). \bibinfo{publisher}{Springer Berlin Heidelberg}, \bibinfo{address}{Berlin, Heidelberg}, \bibinfo{pages}{184--199}.
\newblock
\showISBNx{978-3-540-48308-3}


\bibitem[Chen et~al\mbox{.}(2025)]%
        {chen2025adaptive}
\bibfield{author}{\bibinfo{person}{Yuxuan Chen}, \bibinfo{person}{Rongpeng Li}, \bibinfo{person}{Xiaoxue Yu}, \bibinfo{person}{Zhifeng Zhao}, {and} \bibinfo{person}{Honggang Zhang}.} \bibinfo{year}{2025}\natexlab{}.
\newblock \showarticletitle{Adaptive layer splitting for wireless large language model inference in edge computing: a model-based reinforcement learning approach}.
\newblock \bibinfo{journal}{\emph{Front. Inf. Technol. Electron. Eng.}} \bibinfo{volume}{26}, \bibinfo{number}{2} (\bibinfo{year}{2025}), \bibinfo{pages}{278--292}.
\newblock


\bibitem[Ha and Schmidhuber(2018)]%
        {ha2018world}
\bibfield{author}{\bibinfo{person}{David Ha} {and} \bibinfo{person}{J{\"u}rgen Schmidhuber}.} \bibinfo{year}{2018}\natexlab{}.
\newblock \showarticletitle{World models}.
\newblock \bibinfo{journal}{\emph{arXiv preprint arXiv:1803.10122}} \bibinfo{volume}{2}, \bibinfo{number}{3} (\bibinfo{year}{2018}).
\newblock


\bibitem[He et~al\mbox{.}(2024)]%
        {he2024large}
\bibfield{author}{\bibinfo{person}{Ying He}, \bibinfo{person}{Jingcheng Fang}, \bibinfo{person}{F~Richard Yu}, {and} \bibinfo{person}{Victor~C Leung}.} \bibinfo{year}{2024}\natexlab{}.
\newblock \showarticletitle{Large language models (LLMs) inference offloading and resource allocation in cloud-edge computing: An active inference approach}.
\newblock \bibinfo{journal}{\emph{IEEE Transactions on Mobile Computing}} (\bibinfo{year}{2024}).
\newblock


\bibitem[Howard et~al\mbox{.}(2017)]%
        {howard2017mobilenetsefficientconvolutionalneural}
\bibfield{author}{\bibinfo{person}{Andrew~G. Howard} {et~al\mbox{.}}} \bibinfo{year}{2017}\natexlab{}.
\newblock \bibinfo{title}{MobileNets: Efficient Convolutional Neural Networks for Mobile Vision Applications}.
\newblock
\newblock
\showeprint{1704.04861}


\bibitem[Hu et~al\mbox{.}(2019)]%
        {8737614}
\bibfield{author}{\bibinfo{person}{Chuang Hu}, \bibinfo{person}{Wei Bao}, \bibinfo{person}{Dan Wang}, {and} \bibinfo{person}{Fengming Liu}.} \bibinfo{year}{2019}\natexlab{}.
\newblock \showarticletitle{Dynamic Adaptive DNN Surgery for Inference Acceleration on the Edge}. In \bibinfo{booktitle}{\emph{Proc. IEEE Int. Conf. Comput. Commun. (INFOCOM)}}. \bibinfo{pages}{1423--1431}.
\newblock


\bibitem[Hu et~al\mbox{.}(2022)]%
        {9996638}
\bibfield{author}{\bibinfo{person}{Yang Hu}, \bibinfo{person}{Connor Imes}, \bibinfo{person}{Xuanang Zhao}, \bibinfo{person}{Souvik Kundu}, \bibinfo{person}{Peter~A. Beerel}, \bibinfo{person}{Stephen~P. Crago}, {and} \bibinfo{person}{John~Paul Walters}.} \bibinfo{year}{2022}\natexlab{}.
\newblock \showarticletitle{PipeEdge: Pipeline Parallelism for Large-Scale Model Inference on Heterogeneous Edge Devices}. In \bibinfo{booktitle}{\emph{2022 25th Euromicro Conference on Digital System Design (DSD)}}. \bibinfo{pages}{298--307}.
\newblock
\urldef\tempurl%
\url{https://doi.org/10.1109/DSD57027.2022.00048}
\showDOI{\tempurl}


\bibitem[Jia et~al\mbox{.}(2022)]%
        {10.1145/3498361.3538932}
\bibfield{author}{\bibinfo{person}{Fucheng Jia} {et~al\mbox{.}}} \bibinfo{year}{2022}\natexlab{}.
\newblock \showarticletitle{CoDL: efficient CPU-GPU co-execution for deep learning inference on mobile devices}. In \bibinfo{booktitle}{\emph{Proceedings of the 20th Annual International Conference on Mobile Systems, Applications and Services}} (Portland, Oregon) \emph{(\bibinfo{series}{MobiSys '22})}. \bibinfo{publisher}{Association for Computing Machinery}, \bibinfo{address}{New York, NY, USA}, \bibinfo{pages}{209–221}.
\newblock
\showISBNx{9781450391856}


\bibitem[Jin and Wu(2024)]%
        {jin2024collm}
\bibfield{author}{\bibinfo{person}{Hongpeng Jin} {and} \bibinfo{person}{Yanzhao Wu}.} \bibinfo{year}{2024}\natexlab{}.
\newblock \showarticletitle{Ce-collm: Efficient and adaptive large language models through cloud-edge collaboration}.
\newblock \bibinfo{journal}{\emph{arXiv preprint arXiv:2411.02829}} (\bibinfo{year}{2024}).
\newblock


\bibitem[Kafetzis et~al\mbox{.}(2025)]%
        {kafetzis2025large}
\bibfield{author}{\bibinfo{person}{Dimitrios Kafetzis}, \bibinfo{person}{Ramin Khalili}, {and} \bibinfo{person}{Iordanis Koutsopoulos}.} \bibinfo{year}{2025}\natexlab{}.
\newblock \showarticletitle{Large Language Model partitioning for low-latency inference at the edge}.
\newblock \bibinfo{journal}{\emph{arXiv preprint arXiv:2505.02533}} (\bibinfo{year}{2025}).
\newblock


\bibitem[Kang et~al\mbox{.}(2017)]%
        {10.1145/3037697.3037698}
\bibfield{author}{\bibinfo{person}{Yiping Kang} {et~al\mbox{.}}} \bibinfo{year}{2017}\natexlab{}.
\newblock \showarticletitle{Neurosurgeon: Collaborative Intelligence Between the Cloud and Mobile Edge}. In \bibinfo{booktitle}{\emph{Proceedings of the Twenty-Second International Conference on Architectural Support for Programming Languages and Operating Systems}} (Xi'an, China) \emph{(\bibinfo{series}{ASPLOS '17})}. \bibinfo{publisher}{Association for Computing Machinery}, \bibinfo{address}{New York, NY, USA}, \bibinfo{pages}{615–629}.
\newblock
\showISBNx{9781450344654}


\bibitem[Konrad et~al\mbox{.}(2001)]%
        {konrad2001markov}
\bibfield{author}{\bibinfo{person}{Almudena Konrad}, \bibinfo{person}{Ben~Y Zhao}, \bibinfo{person}{Anthony~D Joseph}, {and} \bibinfo{person}{Reiner Ludwig}.} \bibinfo{year}{2001}\natexlab{}.
\newblock \showarticletitle{A Markov-based channel model algorithm for wireless networks}. In \bibinfo{booktitle}{\emph{Proceedings of the 4th ACM international workshop on Modeling, analysis and simulation of wireless and mobile systems}}. \bibinfo{pages}{28--36}.
\newblock


\bibitem[Laskaridis et~al\mbox{.}(2020)]%
        {Laskaridis_2020}
\bibfield{author}{\bibinfo{person}{Stefanos Laskaridis}, \bibinfo{person}{Stylianos~I. Venieris}, \bibinfo{person}{Mario Almeida}, \bibinfo{person}{Ilias Leontiadis}, {and} \bibinfo{person}{Nicholas~D. Lane}.} \bibinfo{year}{2020}\natexlab{}.
\newblock \showarticletitle{SPINN: synergistic progressive inference of neural networks over device and cloud}. In \bibinfo{booktitle}{\emph{Proceedings of the 26th Annual International Conference on Mobile Computing and Networking}} \emph{(\bibinfo{series}{MobiCom ’20})}. \bibinfo{publisher}{ACM}, \bibinfo{pages}{1–15}.
\newblock


\bibitem[Li et~al\mbox{.}(2019)]%
        {li2019edgeaiondemandaccelerating}
\bibfield{author}{\bibinfo{person}{En Li}, \bibinfo{person}{Liekang Zeng}, \bibinfo{person}{Zhi Zhou}, {and} \bibinfo{person}{Xu Chen}.} \bibinfo{year}{2019}\natexlab{}.
\newblock \bibinfo{title}{Edge AI: On-Demand Accelerating Deep Neural Network Inference via Edge Computing}.
\newblock
\newblock
\showeprint[arxiv]{1910.05316}~[cs.NI]
\urldef\tempurl%
\url{https://arxiv.org/abs/1910.05316}
\showURL{%
\tempurl}


\bibitem[Li et~al\mbox{.}(2025)]%
        {li2025collaborative}
\bibfield{author}{\bibinfo{person}{Senyao Li} {et~al\mbox{.}}} \bibinfo{year}{2025}\natexlab{}.
\newblock \showarticletitle{Collaborative Inference and Learning between Edge SLMs and Cloud LLMs: A Survey of Algorithms, Execution, and Open Challenges}.
\newblock \bibinfo{journal}{\emph{arXiv preprint arXiv:2507.16731}} (\bibinfo{year}{2025}).
\newblock


\bibitem[Malka et~al\mbox{.}(2022)]%
        {malka2022decentralizedlowlatencycollaborativeinference}
\bibfield{author}{\bibinfo{person}{May Malka}, \bibinfo{person}{Erez Farhan}, \bibinfo{person}{Hai Morgenstern}, {and} \bibinfo{person}{Nir Shlezinger}.} \bibinfo{year}{2022}\natexlab{}.
\newblock \bibinfo{title}{Decentralized Low-Latency Collaborative Inference via Ensembles on the Edge}.
\newblock
\newblock
\showeprint[arxiv]{2206.03165}~[cs.LG]
\urldef\tempurl%
\url{https://arxiv.org/abs/2206.03165}
\showURL{%
\tempurl}


\bibitem[Narayan et~al\mbox{.}(2025)]%
        {narayan2025minions}
\bibfield{author}{\bibinfo{person}{Avanika Narayan}, \bibinfo{person}{Dan Biderman}, \bibinfo{person}{Sabri Eyuboglu}, \bibinfo{person}{Avner May}, \bibinfo{person}{Scott Linderman}, \bibinfo{person}{James Zou}, {and} \bibinfo{person}{Christopher Re}.} \bibinfo{year}{2025}\natexlab{}.
\newblock \showarticletitle{Minions: Cost-efficient collaboration between on-device and cloud language models}.
\newblock \bibinfo{journal}{\emph{arXiv preprint arXiv:2502.15964}} (\bibinfo{year}{2025}).
\newblock


\bibitem[Noh et~al\mbox{.}(2025)]%
        {noh2025adaptive}
\bibfield{author}{\bibinfo{person}{Hyeonho Noh}, \bibinfo{person}{Byonghyo Shim}, {and} \bibinfo{person}{Hyun~Jong Yang}.} \bibinfo{year}{2025}\natexlab{}.
\newblock \showarticletitle{Adaptive resource allocation optimization using large language models in dynamic wireless environments}.
\newblock \bibinfo{journal}{\emph{IEEE Trans. Veh. Technol.}} (\bibinfo{year}{2025}).
\newblock


\bibitem[Oustad and Others(2025)]%
        {10994299}
\bibfield{author}{\bibinfo{person}{Elyas Oustad} {and} \bibinfo{person}{Others}.} \bibinfo{year}{2025}\natexlab{}.
\newblock \showarticletitle{DIST: Distributed Learning-Based Energy-Efficient and Reliable Task Scheduling and Resource Allocation in Fog Computing}.
\newblock \bibinfo{journal}{\emph{IEEE Trans. Serv. Comput.}} \bibinfo{volume}{18}, \bibinfo{number}{3} (\bibinfo{year}{2025}), \bibinfo{pages}{1336--1351}.
\newblock


\bibitem[Pan et~al\mbox{.}(2025)]%
        {pan2025instattention}
\bibfield{author}{\bibinfo{person}{Xiurui Pan} {et~al\mbox{.}}} \bibinfo{year}{2025}\natexlab{}.
\newblock \showarticletitle{InstAttention: In-Storage Attention Offloading for Cost-Effective Long-Context LLM Inference}. In \bibinfo{booktitle}{\emph{Proc. IEEE Int. Symp. High Perform. Comput. Archit. (HPCA)}}. IEEE, \bibinfo{pages}{1510--1525}.
\newblock


\bibitem[Pham et~al\mbox{.}(2018)]%
        {8457785}
\bibfield{author}{\bibinfo{person}{Thanh-Phuong Pham}, \bibinfo{person}{Sasko Ristov}, {and} \bibinfo{person}{Thomas Fahringer}.} \bibinfo{year}{2018}\natexlab{}.
\newblock \showarticletitle{Performance and Behavior Characterization of Amazon EC2 Spot Instances}. In \bibinfo{booktitle}{\emph{2018 IEEE 11th International Conference on Cloud Computing (CLOUD)}}. \bibinfo{pages}{73--81}.
\newblock
\urldef\tempurl%
\url{https://doi.org/10.1109/CLOUD.2018.00017}
\showDOI{\tempurl}


\bibitem[Radford et~al\mbox{.}(2019)]%
        {radford2019language}
\bibfield{author}{\bibinfo{person}{Alec Radford}, \bibinfo{person}{Jeffrey Wu}, \bibinfo{person}{Rewon Child}, \bibinfo{person}{David Luan}, \bibinfo{person}{Dario Amodei}, \bibinfo{person}{Ilya Sutskever}, {et~al\mbox{.}}} \bibinfo{year}{2019}\natexlab{}.
\newblock \showarticletitle{Language models are unsupervised multitask learners}.
\newblock \bibinfo{journal}{\emph{OpenAI blog}} \bibinfo{volume}{1}, \bibinfo{number}{8} (\bibinfo{year}{2019}), \bibinfo{pages}{9}.
\newblock


\bibitem[Schulman et~al\mbox{.}(2017)]%
        {schulman2017proximal}
\bibfield{author}{\bibinfo{person}{John Schulman}, \bibinfo{person}{Filip Wolski}, \bibinfo{person}{Prafulla Dhariwal}, \bibinfo{person}{Alec Radford}, {and} \bibinfo{person}{Oleg Klimov}.} \bibinfo{year}{2017}\natexlab{}.
\newblock \showarticletitle{Proximal policy optimization algorithms}.
\newblock \bibinfo{journal}{\emph{arXiv preprint arXiv:1707.06347}} (\bibinfo{year}{2017}).
\newblock


\bibitem[Thoman et~al\mbox{.}(2019)]%
        {10.1007/978-3-030-29400-7_21}
\bibfield{author}{\bibinfo{person}{Peter Thoman}, \bibinfo{person}{Philip Salzmann}, \bibinfo{person}{Biagio Cosenza}, {and} \bibinfo{person}{Thomas Fahringer}.} \bibinfo{year}{2019}\natexlab{}.
\newblock \showarticletitle{Celerity: High-Level C++ for Accelerator Clusters}. In \bibinfo{booktitle}{\emph{Euro-Par 2019: Parallel Processing}}, \bibfield{editor}{\bibinfo{person}{Ramin Yahyapour}} (Ed.). \bibinfo{publisher}{Springer International Publishing}, \bibinfo{address}{Cham}, \bibinfo{pages}{291--303}.
\newblock
\showISBNx{978-3-030-29400-7}


\bibitem[Tian et~al\mbox{.}(2025)]%
        {tian2025clone}
\bibfield{author}{\bibinfo{person}{Chunlin Tian}, \bibinfo{person}{Xinpeng Qin}, \bibinfo{person}{Kahou Tam}, \bibinfo{person}{Li Li}, \bibinfo{person}{Zijian Wang}, \bibinfo{person}{Yuanzhe Zhao}, \bibinfo{person}{Minglei Zhang}, {and} \bibinfo{person}{Chengzhong Xu}.} \bibinfo{year}{2025}\natexlab{}.
\newblock \showarticletitle{CLONE: Customizing LLMs for Efficient Latency-Aware Inference at the Edge}.
\newblock \bibinfo{journal}{\emph{arXiv preprint arXiv:2506.02847}} (\bibinfo{year}{2025}).
\newblock


\bibitem[Ye et~al\mbox{.}(2025)]%
        {ye2025jupiter}
\bibfield{author}{\bibinfo{person}{Shengyuan Ye} {et~al\mbox{.}}} \bibinfo{year}{2025}\natexlab{}.
\newblock \showarticletitle{Jupiter: Fast and resource-efficient collaborative inference of generative llms on edge devices}. In \bibinfo{booktitle}{\emph{Proc. IEEE Int. Conf. Comput. Commun. (INFOCOM)}}. IEEE, \bibinfo{pages}{1--10}.
\newblock


\bibitem[Yuan et~al\mbox{.}(2024)]%
        {yuan2024generative}
\bibfield{author}{\bibinfo{person}{Xingyu Yuan}, \bibinfo{person}{He Li}, \bibinfo{person}{Kaoru Ota}, {and} \bibinfo{person}{Mianxiong Dong}.} \bibinfo{year}{2024}\natexlab{}.
\newblock \showarticletitle{Generative Inference of Large Language Models in Edge Computing: An Energy Efficient Approach}. In \bibinfo{booktitle}{\emph{Proc. Int. Wireless Commun. Mobile Comput. Conf. (IWCMC)}}. \bibinfo{publisher}{IEEE}, \bibinfo{pages}{244--249}.
\newblock


\bibitem[Zhang et~al\mbox{.}(2025)]%
        {zhang2025tensallo}
\bibfield{author}{\bibinfo{person}{Bowen Zhang}, \bibinfo{person}{Junyang Zhang}, \bibinfo{person}{Jiahui Hou}, {and} \bibinfo{person}{Yixin Wang}.} \bibinfo{year}{2025}\natexlab{}.
\newblock \showarticletitle{TensAllo: Adaptive Deployment of LLMs on Resource-Constrained Heterogeneous Edge Devices}. In \bibinfo{booktitle}{\emph{Proc. IEEE Int. Conf. Comput. Commun. (INFOCOM)}}. IEEE, \bibinfo{pages}{1--10}.
\newblock


\bibitem[Zhang et~al\mbox{.}(2024)]%
        {zhang2024edgeshard}
\bibfield{author}{\bibinfo{person}{Mingjin Zhang}, \bibinfo{person}{Xiaoming Shen}, \bibinfo{person}{Jiannong Cao}, \bibinfo{person}{Zeyang Cui}, {and} \bibinfo{person}{Shan Jiang}.} \bibinfo{year}{2024}\natexlab{}.
\newblock \showarticletitle{Edgeshard: Efficient llm inference via collaborative edge computing}.
\newblock \bibinfo{journal}{\emph{IEEE Internet of Things Journal}} (\bibinfo{year}{2024}).
\newblock


\bibitem[Zheng et~al\mbox{.}(2023)]%
        {zheng2023lmsys}
\bibfield{author}{\bibinfo{person}{Lianmin Zheng} {et~al\mbox{.}}} \bibinfo{year}{2023}\natexlab{}.
\newblock \showarticletitle{Lmsys-chat-1m: A large-scale real-world llm conversation dataset}.
\newblock \bibinfo{journal}{\emph{arXiv preprint arXiv:2309.11998}} (\bibinfo{year}{2023}).
\newblock


\bibitem[Zhu and Yang(2025)]%
        {zhu2025lslm}
\bibfield{author}{\bibinfo{person}{Pengyan Zhu} {and} \bibinfo{person}{Tingting Yang}.} \bibinfo{year}{2025}\natexlab{}.
\newblock \showarticletitle{CE-LSLM: Efficient Large-Small Language Model Inference and Communication via Cloud-Edge Collaboration}.
\newblock \bibinfo{journal}{\emph{arXiv preprint arXiv:2505.14085}} (\bibinfo{year}{2025}).
\newblock


\end{thebibliography}

\end{document}